\documentclass[letterpaper,journal]{IEEEtran}
\usepackage{amsmath,amsfonts}
\usepackage{algorithm}
\usepackage{algpseudocode}
\usepackage{array}
\usepackage{booktabs}
\usepackage[caption=false,font=normalsize,labelfont=sf,textfont=sf]{subfig}
\usepackage{textcomp}
\usepackage{stfloats}
\usepackage{url}
\usepackage{verbatim}
\usepackage{graphicx}
\usepackage{cite}

\begin{document}

\title{A 7-DOF PAM-Driven Anthropomorphic Arm with Load-Adaptive, History-Aware Neural Control}

\author{Qiyao Wang$^{1}$$^{\dagger}$, Zhonghan Lin$^{1}$$^{\dagger}$, Yinglei Zhu$^{1}$, Qi Shao$^{1}$, Xin-Jun Liu$^{1}$ and Huichan Zhao$^{1}$$^{*}$%
\thanks{$^{\dagger}$Contributed equally to this work.}%
\thanks{$^{1}$Department of Mechanical Engineering, State Key Laboratory of Tribology in Advanced Equipment, Beijing Key Laboratory of Transformative High-end Manufacturing Equipment and Technology, Tsinghua University, Beijing 100084, China.}%
\thanks{$^{*}$Corresponding Author}%
}

\maketitle

\begin{abstract}
  Pneumatic artificial muscle (PAM)-driven manipulators combine rigid-link kinematics with compliant soft actuation, but control is difficult because elongation depends on load, motion history, specimen variation, and long-term drift. We present a 7-DOF anthropomorphic pneumatic arm reproducing the shoulder--elbow--wrist chain and DANC, a Dual-Channel Adaptive Neural Controller for load-adaptive, history-aware PAM control. Twenty-two custom-fabricated multi-fiber PAMs arranged in antagonistic groups across three spherical joints drive the arm. Actuator and joint characterization links pressure--load--elongation behavior to torque, stiffness, and workspace, showing how actuator-level uncertainty accumulates in full-arm motion.
  
  DANC combines load-conditioned pressure inference using elongation history and target preview, adaptive mass-equivalent load estimation, and rapid stiffness-ratio calibration for cross-specimen transfer without retraining. Single-PAM experiments demonstrate tracking under 0--8 kg and changing loads. At 5 kg, 10-mm-step rise and 5\% settling times are 0.17 and 0.62 s. Its second-order-reference-normalized time ratios are lowest among comparable published cases; ablations support its design. The 3-DOF joint tracks star-shaped paths with a best RMSE of 6.7 mm; the full arm executes coordinated 7-DOF motion with maximum sampled end-effector errors of 34.9 mm and 2.74$^\circ$. Additional tests cover one-PAM tracking after 12 months, arm operation after removing one PAM, and drag-following interaction.
\end{abstract}



\section{Introduction}
\IEEEPARstart{P}{neumatic} artificial muscles (PAMs) provide a promising actuation principle for robots that must move near and interact with people. They are lightweight, compliant, and capable of producing large forces relative to their mass, making them well suited for physical interaction and biomimetic motion \cite{daerden2002pneumatic,zhang2019robotic}. Particularly, they are valuable for soft--rigid hybrid manipulators, where an articulated skeleton provides a well-defined kinematic structure, and the muscle-like actuation supplies compliance during physical interactions \cite{rus2015design,bicchi2004fast}. Recent musculoskeletal platforms built around artificial muscles of different physical principles---from thin braided McKibben bundles to electrohydraulic actuators---underline the sustained effort and prospect in this architecture \cite{na2023braided,buchner2024electrohydraulic}.

The same properties also make PAMs difficult to control. In the fiber-reinforced extensile PAMs used here, elongation depends nonlinearly on pressure and load \cite{chou1996measurement}, with further contributions from friction-dominated quasi-static hysteresis, rate-dependent hysteresis arising from viscoelasticity and pneumatic dynamics, creep, and specimen variation
 \cite{davis2006braid,reynolds2003modeling,kalita2022review}. The pressure required for a target elongation is therefore not a fixed map: it shifts with external load and intrinsic actuation history. In an anthropomorphic arm built from such PAMs, this actuator-level uncertainty is further amplified by the antagonistic routing around the multi-DOF joints, where small elongation errors may accumulate into considerable torque imbalance, posture error, and stiffness variation.

\begin{figure*}[!t]
\centering
\includegraphics[width=\textwidth]{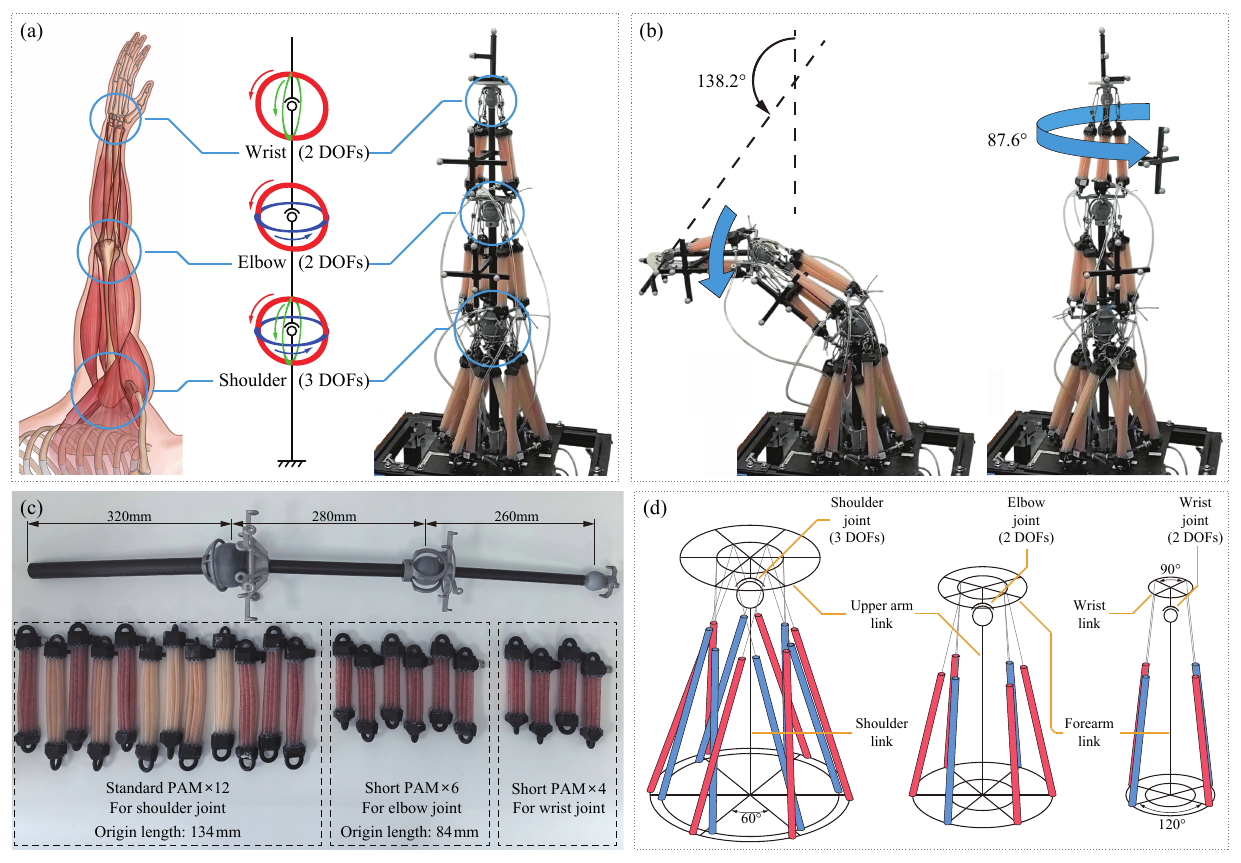}
\caption{Overview of the 7-DOF anthropomorphic pneumatic manipulator. (a) Biomimetic DOF allocation of the shoulder, elbow, and wrist compared with the human upper limb and the assembled arm. (b) Measured bending and torsional motion ranges of the arm. (c) Link dimensions and PAM distribution, including standard PAMs for the shoulder and short PAMs for the elbow and wrist. (d) Schematic of the spherical joints, rigid links, and antagonistic PAM routing around the shoulder, elbow, and wrist modules.}
\label{fig:intro}
\end{figure*}

This paper addresses this problem on the 7-DOF anthropomorphic pneumatic manipulator shown in Fig.~\ref{fig:intro}(a). The arm reproduces the shoulder--elbow--wrist structure of the human upper limb, with a 3-DOF shoulder, 2-DOF elbow, and 2-DOF wrist, and is driven by 22 custom-fabricated PAMs, each composed of 12 muscle fibers, arranged in antagonistic groups (Fig.~\ref{fig:intro}(c)). Compared to prior PAM arms, the platform introduces two mechanical advances: it extends the single-joint soft-joint concept proposed in \cite{lin2022anthropomorphic} to a full three-joint, human-scale limb with crossed antagonistic routing, and it adapts the fiber-reinforcement principle of \cite{hawkes2016design} into a rapid small-batch process that yields the joint-specific standard- and short-length actuators required by this dense layout. This musculoskeletal platform not only serves as a paradigm for full‑DOF anthropomorphic arm design, but also provides a sufficiently rich experimental foundation for load‑adaptive, history‑aware PAM control.

To address the control difficulty, we develop a Dual-Channel Adaptive Neural Controller (DANC) for calculating PAM's driving pressure under unknown and changing loads. A Pressure Inference Channel (PIC) learns the inverse pressure mappings from temporally extended motion data under several predefined loads and combines them into a continuous pressure-load function. This is a learned counterpart of multi-model and gain-scheduled adaptive control \cite{narendra1997adaptive}, except that the scheduling variable---the effective load---cannot be measured on the arm. A Load Estimation Channel (LEC) therefore estimates the real-time load online by combining dynamic neural inference, Kalman filtering, and static compensation. Substituting the estimated load into the PIC profile yields a pressure command that adapts online to payload variation and interaction disturbances, while allowing specimen variation and drift to be absorbed as load-estimation bias rather than elongation-tracking error.

The paper makes three major contributions:
\begin{itemize}
    \item We integrate a human-scale 7-DOF anthropomorphic PAM arm with 22 multi-fiber extensile PAMs arranged in antagonistic groups across the shoulder, elbow, and wrist, supported by a rapid small-batch fabrication process for joint-specific standard- and short-length actuators.
    \item We characterize the PAMs and joints to identify the load- and history-dependent actuator behavior and posture-dependent joint mechanics that guide controller design and experimental interpretation.
    \item We propose DANC, which combines load-conditioned pressure inference with online load estimation for history-aware PAM regulation, together with a rapid stiffness-ratio calibration that transfers the trained controller across PAM specimens without retraining, validated from single actuators to the full arm.
\end{itemize}

\section{Related Work}
\label{sec:related}
\subsection{Soft--Rigid and Musculoskeletal PAM Manipulators}
Soft--rigid coupled manipulators retain a skeleton-like kinematic chain while introducing compliant actuation or compliant joints, combining positioning capability with compliant physical interaction \cite{rus2015design,bicchi2004fast,zuo2026realtimecompliancepositioncontrol}. PAMs are well matched to this goal because they provide lightweight, compliant actuation with large stress, large strain, and fast response \cite{daerden2002pneumatic,zhang2019robotic,dong2025design}.

PAM-driven anthropomorphic arms have demonstrated the feasibility of compliant upper-limb machines. Tuijthof and Herder designed a 4-DOF arm with gravity compensation and open-loop stiffness control \cite{tuijthof2000design}. Tondu \emph{et al.} presented a seven-revolute-joint humanoid arm driven by antagonistic McKibben pairs \cite{tondu2005seven}. Ohta \emph{et al.} developed a lightweight 7-DOF arm combining rigid links, custom McKibben muscles, and inflatable sleeves \cite{ohta2018design}, while Bruder, Wood, and co-workers increased soft-arm payload capacity through localized stiffening \cite{bruder2023increasing}. Musculoskeletal platforms go further by arranging many compliant actuators around biologically inspired joints, including upper limbs for embodied-intelligence studies \cite{hosoda2012anthropomorphic}, multifilament muscle bundles for human-scale limbs \cite{na2023braided}, soft-joint upper-limb modules \cite{lin2022anthropomorphic}, and learning-based dynamic control for soft robotic manipulators \cite{thuruthel2018model}; beyond pneumatics, electrohydraulic musculoskeletal systems illustrate the breadth of this paradigm \cite{buchner2024electrohydraulic}.

Together, these studies show the promise of PAM-based compliant upper-limb robots, but reliable actuator-level regulation remains difficult in multi-joint systems. When many PAMs act through coupled joints, small actuator-level errors can accumulate into joint torque imbalance and whole-arm posture error. A major source of such errors is load variation: the pressure required to reach a target elongation changes with posture, payload, and contact. This work therefore focuses on load-adaptive PAM regulation as the actuator-level basis for controlling a 7-DOF anthropomorphic arm.

\subsection{PAM Modeling and Model-Based Control}
PAM modeling has been studied extensively, and structured overviews of soft-robot modeling place these actuator models in a broader context \cite{armanini2023soft}. Classical geometric and phenomenological models remain common references for McKibben-type actuator analysis \cite{chou1996measurement,tondu2000modeling}. Dynamic models based on viscoelastic elements and reviews of McKibben modeling assumptions further clarify the actuator's rate-dependent behavior and model limitations \cite{reynolds2003modeling,tondu2014robust}. These models provide physical insight, but their parameters are sensitive to actuator construction, friction, aging, and load.

Model-based controllers often combine the above-mentioned models with feedback, adaptation, or hysteresis compensation. Representative methods include adaptive tracking \cite{lilly2003adaptive}, Maxwell-slip and Prandtl--Ishlinskii hysteresis compensation \cite{minh2010cascade,vo2010new,xie2018modeling}, constrained optimal control \cite{andrikopoulos2013piecewise}, robust adaptive and proxy-based sliding-mode control \cite{zhu2008tmech,huang2022adaptive}, disturbance-observer methods \cite{wu2013nonlinear}, port-Hamiltonian control \cite{tanaka2025position}, and active model-based compensation \cite{zhang2016active,qin2022active}. More recent hybrid schemes use neural networks as online hysteresis compensators whose weights are estimated in real time by Kalman-type filters \cite{zhang2024active}.

Learning-based model structures can reduce reliance on hand-crafted dynamics; for example, deep Lagrangian networks have recently been combined with reinforcement learning for PAM modeling and control \cite{wang2025dynamic}. Nevertheless, these methods are typically evaluated in single-actuator, antagonistic-pair, or low-DOF settings, where the relevant load range, hysteresis, and mechanism dynamics can be identified for a limited set of actuators. In a dense anthropomorphic PAM arm, the same characterization would have to be maintained for many custom PAMs over changing postures, payloads, interactions, and material drift. This motivates a data-driven inverse control strategy that learns the pressure command directly from measured PAM behavior.

\subsection{Data-Driven PAM Control and Load Adaptation}
Data-driven methods have become important in PAM and soft-robot control because they approximate the nonlinear dynamics that are hard to model analytically. Recurrent and reservoir networks capture temporal PAM dynamics and support online adaptation under nonlinearities and disturbances \cite{ahn2009design,xing2012modeling}, a line recently extended to pneumatic physical reservoir computing for feedforward hysteresis compensation \cite{shen2024control}. Other controllers embed neural, adaptive, or hybrid elements inside feedback loops \cite{robinson2015nonlinear,ba2016integrated,son2017novel,chavoshian2020hybrid}. Learned dynamics have also been used in predictive control for PAM actuators, exoskeletons, and broader soft-robot systems \cite{huang2018echo,cao2020neural,bruder2020data}, while recent work explores whole-body learned control \cite{chen2025hysteresis} and simulation-based control of PAM-driven robots \cite{zhu2025equimus}.

Many of these controllers handle load variation implicitly, through online adaptation, robust feedback, or disturbance rejection, rather than representing load as an explicit actuator state; alternatively, load can be measured directly by embedding force or displacement sensors in the actuator \cite{wirekoh2019sensorized}. For a PAM-driven arm, an explicit yet sensor-free load state is attractive: it determines the pressure required for a target elongation, reflects changes in payload and contact, and absorbs specimen variation or drift that appears as an effective load change---without adding hardware to every single actuator.

In this work, our controller realizes this state by learning the required pressure across load conditions---in the spirit of multi-model adaptive control \cite{narendra1997adaptive} but with learned local models---and estimating the unmeasured load online to condition that command, which also serves as a reusable cue for drag-following interaction.

\begin{figure*}[!t]
\centering
\includegraphics[width=\textwidth]{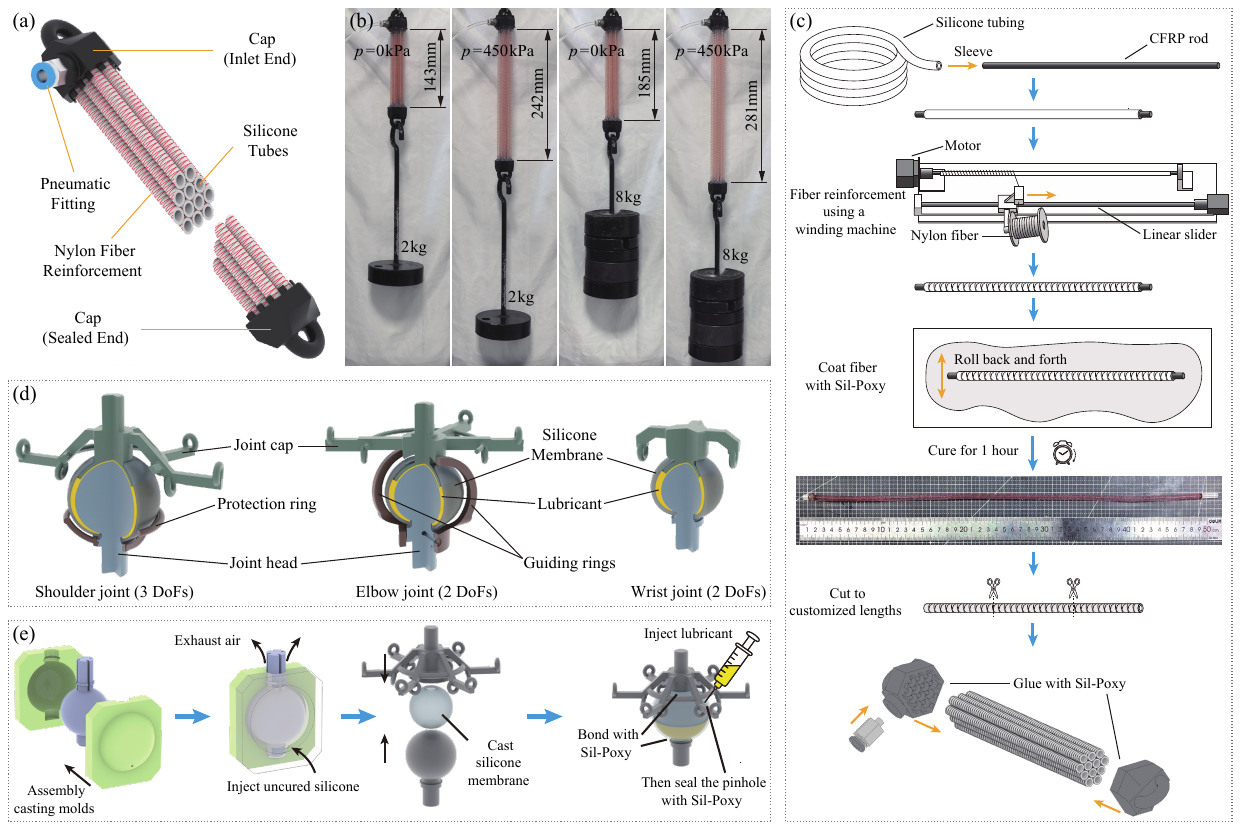}
\caption{Fabrication of the pneumatic artificial muscles and custom spherical joints. (a) Structure of a PAM, including parallel muscle fibers, end caps, and pneumatic fitting. (b) Elongation of short and standard PAMs under representative pressures and loads. (c) PAM fabrication process, including tube sleeving, fiber winding, Sil-Poxy coating, curing, cutting, and end-cap bonding. (d) CAD models of the shoulder, elbow, and wrist spherical joints with joint caps, protection rings, guiding rings, lubricant, and joint heads. (e) Casting and assembly process for the silicone membrane and lubricated spherical joint.}
\label{fig:fabrication}
\end{figure*}

\section{System Design and Fabrication}
\label{sec:system}
\subsection{Anthropomorphic Architecture}
The manipulator is designed as a shoulder--elbow--wrist chain with human-scale proportions. As shown in Fig.~\ref{fig:intro}(a), its DOF allocation follows the human upper limb in both number and direction: the shoulder provides three rotational DOFs, the elbow provides one bending DOF and one forearm-twisting DOF, and the wrist provides two orthogonal bending DOFs. The nominal link lengths are 320, 280, and 260 mm, respectively, as shown in Fig.~\ref{fig:intro}(c). 

Each joint is implemented as a biomimetic spherical module, extending the soft-joint concept in \cite{lin2022anthropomorphic} to the present shoulder--elbow--wrist arm. A spherical (ball-and-socket) interface is adopted because it offers omnidirectional, nearly isotropic rotation, avoiding the direction-dependent properties and singularities of stacked single-axis joints. PAMs are routed around the spherical surface in antagonistic groups, so an imbalance between the forces generated by opposing PAMs produces joint rotation, while co-activation changes the local restoring torque and stiffness. The complete arm uses 22 PAMs: 12 standard PAMs at the shoulder, 6 short PAMs at the elbow, and 4 short PAMs at the wrist. Each PAM contains 12 muscle fibers, giving 264 fibers in total; the original fiber lengths are 134 mm for the standard PAMs and 84 mm for the short PAMs. This dense, multi-fiber arrangement increases pressure-allocation complexity but supports compliance modulation and redundancy across the arm.

The PAM routing pattern is shown in Fig.~\ref{fig:intro}(d). The PAMs are not placed as simple circumferential loops; instead, opposing PAMs are arranged in crossed paths between link-side anchoring points around the spherical joints. This crossed antagonistic layout allows the tensile forces generated by the PAMs to produce torques about both the bending and twisting axes, enabling each module to realize its intended DOFs. The measured motion range of the assembled arm is summarized in Fig.~\ref{fig:intro}(b): the arm reaches a bending range of $138.2^\circ$ and a torsional range of $87.6^\circ$. The routing density is highest at the shoulder because it carries the largest equivalent load and has the largest number of DOFs.

\subsection{PAM Design and Small-Batch Fabrication}
The PAM design is shown in Fig.~\ref{fig:fabrication}(a). Each PAM consists of 12 parallel muscle fibers connected by a pair of shared end caps, a pneumatic fitting, and mechanical attachment features. Bundling the fibers in parallel adds their cross-sectional areas, scaling up the axial force at a given pressure while keeping each individual fiber thin and compliant. Each fiber uses a silicone tube with a 4 mm inner diameter, a 6 mm outer diameter, and a Shore A hardness of 40. The nylon reinforcement is wound nearly circumferentially, restricting radial expansion so that internal pressure produces axial elongation rather than McKibben-type contraction. Fig.~\ref{fig:fabrication}(b) shows the deformation range and payload-lifting capability of the standard and short PAMs.

We propose a rapid small-batch fabrication method for these multi-fiber PAMs, as shown in Fig.~\ref{fig:fabrication}(c). A silicone tube is sleeved over a carbon-fiber-reinforced polymer rod, and a motorized winding stage wraps the nylon fiber with controlled pitch and repeatable tension. The reinforced tube is coated with Sil-Poxy and rolled back and forth to distribute the adhesive uniformly, then cut to length and bonded into common end caps after curing. Excluding the Sil-Poxy curing interval, one operator can fabricate approximately three standard PAMs or four short PAMs per hour.

\subsection{Biomimetic Spherical Joint Design and Fabrication}
The spherical-joint design is shown in Fig.~\ref{fig:fabrication}(d). Each joint uses a joint head, a joint cap, and an outer silicone membrane, following the biomimetic soft-joint idea in \cite{lin2022anthropomorphic}. The membrane connects the joint head and joint cap and wraps around them to form a compliant interface; lubricant is filled inside the membrane to reduce sliding friction. The shoulder module further includes a protection ring that prevents tendons from cutting into the silicone membrane under repeated loading. The elbow module includes guiding rings that constrain one rotational DOF by guiding the trajectory of the joint cap.

The joint fabrication process is shown in Fig.~\ref{fig:fabrication}(e). The outer membrane is cast from Smooth-On Dragon Skin 30 silicone using 3D-printed casting molds. After curing, the membrane is fitted over the joint head and joint cap, and the assembly is bonded with Sil-Poxy. Finally, lubricant is injected into the membrane, and the pinhole is sealed with Sil-Poxy.

\begin{figure*}[!t]
\centering
\includegraphics[width=\textwidth]{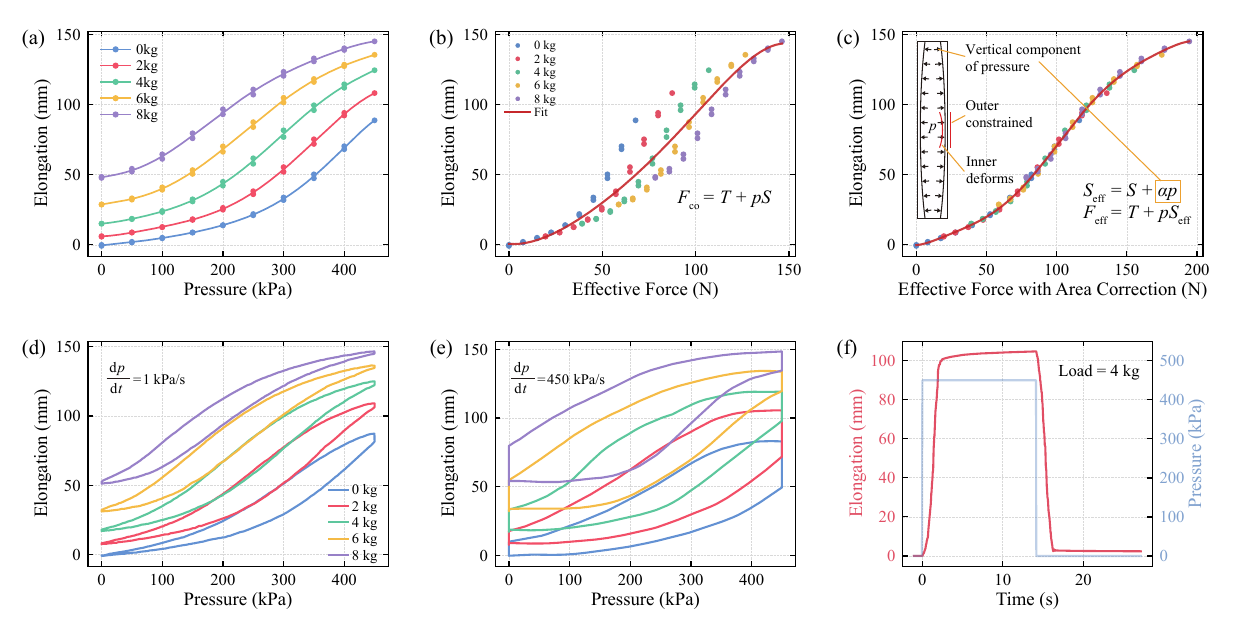}
\caption{Static and dynamic characterization of the fabricated PAMs. (a) Static pressure--elongation curves under 0--8 kg loads. (b) Baseline effective-force fitting with a constant pressure area. (c) Effective-force fitting after pressure-dependent area correction. (d) Friction-dominated quasi-static pressure--elongation hysteresis under a slow pressure sweep. (e) Rate-dependent pressure--elongation hysteresis at a pressure-change rate of 450 kPa/s. (f) Creep response under a 4 kg load and a step pressure input.}
\label{fig:character-muscle}
\end{figure*}

\section{Actuator and Joint Characterization}
\label{sec:characterization}
\subsection{PAM Static Characterization}
The static characterization identifies the actuator behaviors the controller must handle, using step-response steady-state data. For each load, the PAM was driven to pressure values from 0 to 450 kPa, held until the transient response decayed, and then measured. Compared with continuous quasi-static pressure sweeps, this procedure reduces equilibrium-position bias caused by friction and approach-direction effects, thereby isolating the pressure--load--elongation relation used for static compensation. The resulting static pressure--load--elongation data are shown in Fig.~\ref{fig:character-muscle}(a).

After the transient response has settled, the axial equilibrium of a PAM can be approximated by
\begin{equation}
    T - F_{\mathrm{elas}}(x) + pS(p) = 0,
\end{equation}
where $T$ is the axial load, $x$ is elongation, $p$ is pressure, $F_{\mathrm{elas}}(x)$ is the elastomer restoring force, and $S(p)$ is the effective pressure area. If $S$ is assumed constant, the actuator can be approximated with an effective force
\begin{equation}
    F_{\mathrm{eff}} = T + pS,
\end{equation}
and we then fit the elongation with the defined $F_{\mathrm{eff}}$ as a polynomial:
\begin{equation}
    x = \sum_{i=0}^{5} a_i F_{\mathrm{eff}}^i.
\end{equation}
Projecting the data onto $F_{\mathrm{eff}}$ with this constant-area model gives the baseline fit in Fig.~\ref{fig:character-muscle}(b). The data collected under different loads do not collapse onto a single curve, and the fitted polynomial therefore cannot represent the full static relation well. The reason is that the internal silicone tubes and external fiber reinforcement deform with pressure, so the effective pressure area is not constant over the operating range.

We therefore introduce a pressure-dependent area correction:
\begin{equation}
\begin{aligned}
    S_{\mathrm{eff}}(p) &= S_0 + \alpha_p p,\\
    F_{\mathrm{eff}}(T,p) &= T + pS_{\mathrm{eff}}(p)
    = T + S_0p + \alpha_p p^2 .
\end{aligned}
\label{eq:pressure-dependent-effective-force}
\end{equation}
Here $S_0$ denotes the nominal effective pressure coefficient, and $\alpha_p$ captures the first-order pressure dependence of the effective area. In the fitted model, $p$, $T$, and $x$ are expressed in the units of the experimental dataset, so $S_0$ and $\alpha_p$ are treated as effective coefficients rather than independently measured geometric areas.
The static model becomes
\begin{equation}
    x = \sum_{i=0}^{5} a_i \left(T + S_0p + \alpha_p p^2\right)^i .
\label{eq:static-pam-model}
\end{equation}
Because the experimental samples are unevenly distributed over $F_{\mathrm{eff}}$, we identify the parameters using inverse-density weighted least squares. The force axis is divided into bins, and each sample is weighted inversely to the number of samples in its bin:
\begin{equation}
    w_j = \frac{\max_k N_k}{N_{\mathrm{bin}(j)}} .
\end{equation}
Let $\boldsymbol{\theta}=[S_0,\alpha_p,a_0,\ldots,a_5]$ denote the identified parameters. The weighted fitting problem is
\begin{equation}
    \boldsymbol{\theta}^{*} =
    \arg\min_{\boldsymbol{\theta}}
    \sqrt{
    \frac{\sum_{j=1}^{K} w_j
    \left(x_j-\sum_{i=0}^{5}a_iF_{\mathrm{eff},j}^{i}\right)^2}
    {\sum_{j=1}^{K}w_j}
    } .
\end{equation}
Fig.~\ref{fig:character-muscle}(c) shows that the corrected effective-force representation substantially reduces the load-dependent scatter, while the fitted curve remains strongly nonlinear over the operating range. This model therefore serves three purposes: (i) it quantifies the static load--pressure--elongation nonlinearity, (ii) it supports the joint-level static torque and stiffness characterization, and (iii) it provides the static map later used by the load-estimation channel in the controller.

\subsection{PAM Dynamic Characteristics}
The static model captures the settled pressure--load--elongation relation, but the PAM response also depends on path, rate, and time. Fig.~\ref{fig:character-muscle}(d) shows friction-dominated quasi-static pressure--elongation hysteresis under a very slow pressure sweep. At this low rate, rate-dependent effects are limited, and the hysteresis mainly reflects friction and approach-direction-dependent equilibrium.

When the pressure rate increases to 450 kPa/s, Fig.~\ref{fig:character-muscle}(e) shows substantially greater hysteresis than Fig.~\ref{fig:character-muscle}(d), indicating an additional rate-dependent component during rapid inflation and deflation. This component arises from viscoelastic material response and pneumatic dynamics. Fig.~\ref{fig:character-muscle}(f) further shows creep under constant pressure and load. These effects imply that the inverse pressure command cannot depend only on the current desired elongation and load; it must also use recent motion history and the expected direction of upcoming motion.

Beyond these dynamic effects, PAM-to-PAM variation also affects their deployment on a multi-actuator arm. Even within one fabrication batch, small hardness differences and residual manufacturing errors shift the constitutive response between nominally identical PAMs, as shown across six specimens in Fig.~\ref{fig:muscle-control-variation}(a). Because the full arm uses 22 PAMs composed of 264 fibers, the controller must tolerate specimen variation across many actuator units.

\begin{figure*}[!t]
\centering
\includegraphics[width=\textwidth]{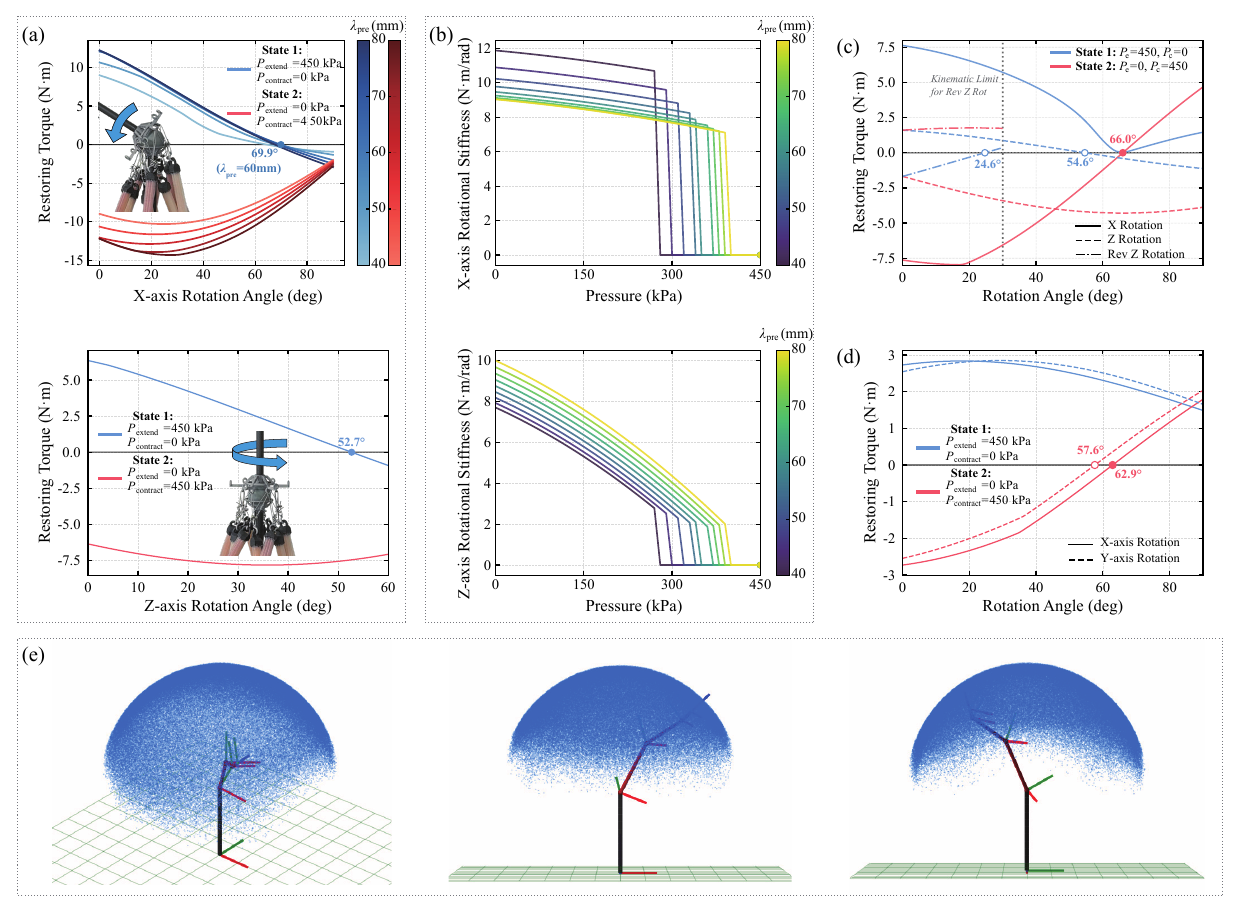}
\caption{Joint-level and whole-arm characterization. (a) Shoulder torque--rotation relations in lateral bending (X direction) and torsion (Z direction). (b) Shoulder variable-stiffness characteristics in the X and Z directions as functions of pressure and pre-elongation. (c) Elbow torque--rotation relations. (d) Wrist torque--rotation relations. (e) Whole-arm workspace visualization.}
\label{fig:character-arm}
\end{figure*}

\subsection{Joint Torque, Stiffness, and Workspace}
Joint-level characterization evaluates how the static PAM model propagates through the redundant spherical module. Consider a joint with $n_{\mathrm{PAM}}$ PAMs. Let $\mathbf{b}_i$ be the fixed-side anchor position of the $i$-th PAM, and let $\mathbf{a}_{i,0}$ be its moving-side anchor position at its neutral posture. For a joint posture $\mathbf{q}$ (a unit quaternion) with rotation matrix $R(\mathbf{q})$, the moving anchor and PAM length are
\begin{equation}
    \mathbf{a}_i(\mathbf{q})=R(\mathbf{q})\mathbf{a}_{i,0},\qquad
    \ell_i(\mathbf{q})=\|\mathbf{a}_i(\mathbf{q})-\mathbf{b}_i\| .
\end{equation}
Here $\ell_i$ is the anchor-to-anchor length, including the PAM, the tendon, and the connection segments. The constant segments cancel when the elongation change is computed relative to the neutral posture.
The cable length is chosen so that the PAM has an initial pre-elongation $\delta_{\mathrm{pre}}$ at its neutral posture $\mathbf{q}_0$. Thus the geometric elongation used by the static PAM model is
\begin{equation}
    x_i(\mathbf{q})=
    \ell_i(\mathbf{q})-\ell_i(\mathbf{q}_0)+\delta_{\mathrm{pre}} .
\end{equation}
The pre-elongation must keep the PAMs tensioned while avoiding excessive stretch,
\begin{equation}
    0 < x_i(\mathbf{q}) < x_{\mathrm{lim}},
    \qquad i=1,\ldots,n ,
\end{equation}
over the intended workspace.

For a given pressure $p_i$ and elongation $x_i$, the static PAM model is inverted numerically to obtain the tendon tension. The effective force $F_i$ is obtained from
\begin{equation}
    \sum_{k=0}^{5}a_k F_i^k-x_i=0 .
\end{equation}
Among the polynomial roots, the physically admissible real root in the operating range is selected. The one-sided tensile nature of the PAM is enforced by
\begin{equation}
    T_i=\max\left(0,\,F_i-S_0p_i-\alpha_p p_i^2\right).
\end{equation}
The vector force applied to the moving link is then
\begin{equation}
    \mathbf{f}_i
    =T_i\frac{\mathbf{b}_i-\mathbf{a}_i(\mathbf{q})}
    {\ell_i(\mathbf{q})},
\end{equation}
and the joint restoring torque about the spherical center is
\begin{equation}
    \boldsymbol{\tau}(\mathbf{q}) =
    \sum_{i=1}^{n_{\mathrm{PAM}}}\mathbf{a}_i(\mathbf{q})\times\mathbf{f}_i .
\end{equation}
The torque--rotation curves in Fig.~\ref{fig:character-arm}(a), (c), and (d) compare two antagonistic pressure states. In State~1, the lengthening-side PAMs are pressurized to 450~kPa and the shortening-side PAMs to 0~kPa, so the zero-torque crossing gives the no-load rotation limit. State~2 reverses this allocation and characterizes load recovery from a deflected posture. A State~2 torque that decreases with rotation indicates locally stable recovery; the shoulder follows this trend at small rotations but loses it at larger angles, exposing a posture-dependent control challenge.

The rotational stiffness is estimated with an energy method because the PAM constitutive relation is nonlinear and the torque derivative is cumbersome to obtain analytically. For a fixed pressure $p_i$, the zero-tension effective force is $F_i^0=S_0p_i+\alpha_p p_i^2$. For the tensioned branch, the potential energy stored in one PAM when its effective force is $F_i$ is
\begin{equation}
    U_i(F_i,p_i)=
    \int_{F_i^0}^{F_i}
    \left(\eta-F_i^0\right)
    \frac{dx}{d\eta}\,d\eta ,
\end{equation}
or, using the fitted polynomial $x=\sum_{k=0}^{5}a_kF^k$,
\begin{equation}
    U_i(F_i,p_i)=
    \left[
    \sum_{k=1}^{5}\frac{k a_k}{k+1}\eta^{k+1}
    -F_i^0\sum_{k=1}^{5}a_k\eta^k
    \right]_{\eta=F_i^0}^{F_i}.
\end{equation}
If the PAM is slack, its tension and incremental stiffness contribution are set to zero. The system potential is $U_{\mathrm{sys}}=\sum_i U_i$. The scalar rotational stiffness about a given axis is obtained by differentiating $U_{\mathrm{sys}}$ twice with respect to the corresponding rotation angle. This gives the pressure- and pre-elongation-dependent stiffness curves in Fig.~\ref{fig:character-arm}(b).

The whole-arm workspace in Fig.~\ref{fig:character-arm}(e) is obtained by sampling feasible shoulder, elbow, and wrist postures and applying the forward kinematics of the three-link chain,
\begin{equation}
    \mathbf{p}_{\mathrm{ee}} =
    R_s\mathbf{d}_1+
    R_sR_e\mathbf{d}_2+
    R_sR_eR_w\mathbf{d}_3 ,
\end{equation}
where $\mathbf{d}_1$, $\mathbf{d}_2$, and $\mathbf{d}_3$ are the nominal link vectors and $R_s$, $R_e$, and $R_w$ are the shoulder, elbow, and wrist rotations. The sampled postures are restricted to the experimentally identified rotation ranges of the shoulder, elbow, and wrist, so Fig.~\ref{fig:character-arm}(e) reports the reachable end-effector region of the assembled arm rather than an unconstrained serial-chain workspace.

Together, the joint-level maps show that the same PAM routing that provides a broad workspace also makes the available restoring torque and stiffness strongly dependent on posture and pressure allocation. Once the links are assembled and held against gravity, the gravitational moment arm of the distal links varies sharply with posture, so the load torque each joint must balance---and hence the load carried by its PAMs---changes strongly across the workspace. The arm is mounted upright, extending above its fixed base in a gravitationally unstable configuration: deviations from the vertical increase the moment arms of the distal links and generate growing destabilizing torques.
This characterization sets the operating context for the controller: actuator-level regulation must remain reliable while the effective load on each PAM changes with joint configuration and external interaction.

\begin{figure*}[!t]
\centering
\includegraphics[width=\textwidth]{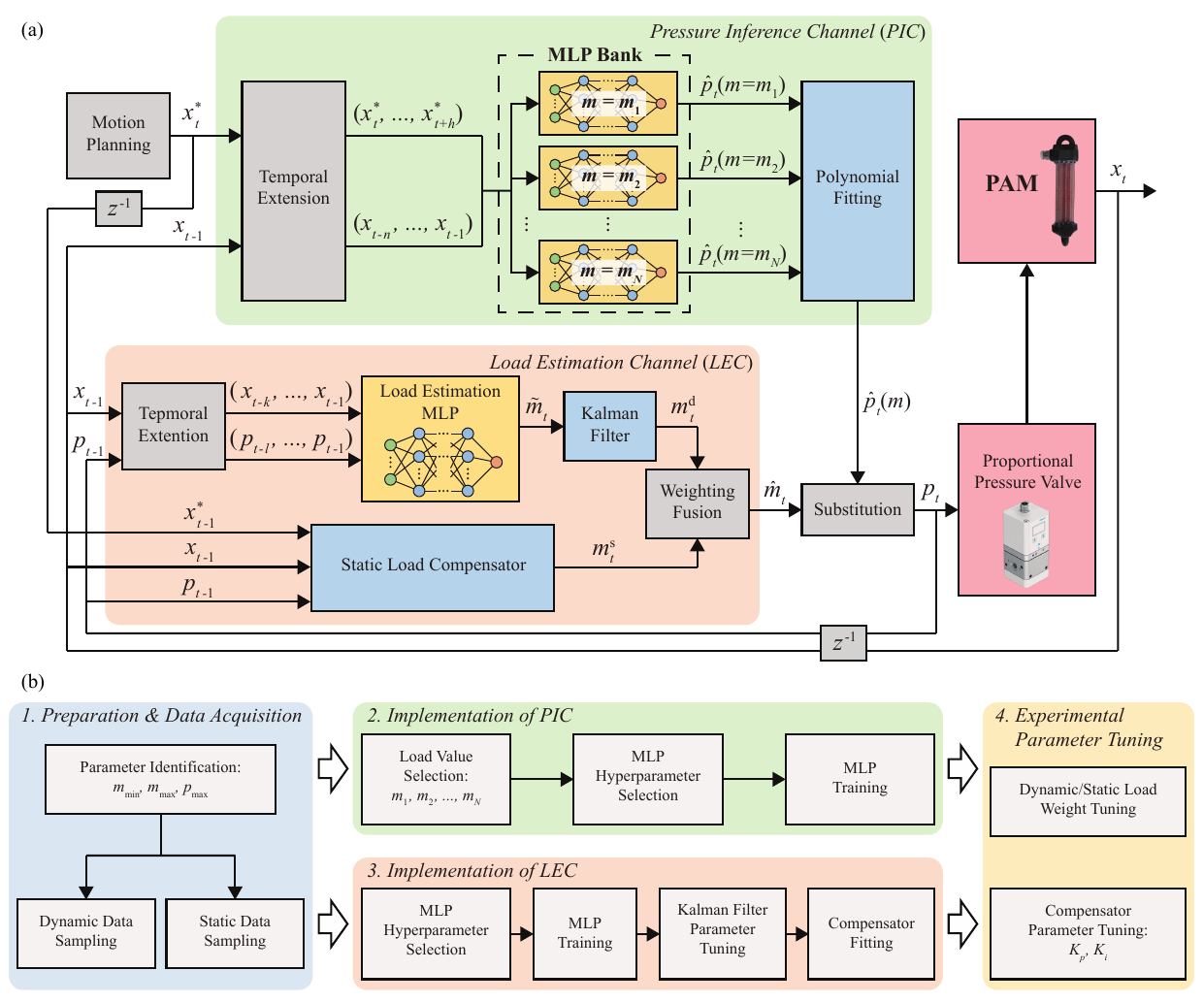}
\caption{Architecture and implementation of the Dual-Channel Adaptive Neural Controller (DANC). (a) Online architecture: the pressure inference channel temporally extends planned and measured elongations, evaluates a load-specific MLP bank, and fits a pressure--load polynomial; the load estimation channel estimates dynamic load from pressure/elongation histories, applies Kalman filtering, fuses the result with static compensation, and substitutes the estimated load to command the valve and PAM. (b) Implementation workflow, including data preparation, pressure-channel training, load-channel training and compensator fitting, and experimental parameter tuning.}
\label{fig:controller}
\end{figure*}

\section{Dual-Channel Adaptive Neural Controller}
\label{sec:control}
The Dual-Channel Adaptive Neural Controller (DANC) is an actuator-level inverse controller that computes the PAM pressure required to realize a desired elongation from the recent motion history and desired-trajectory preview. It uses two coupled channels: the Pressure Inference Channel (PIC) computes the required control pressure as a function of load for the current motion state, while the Load Estimation Channel (LEC) estimates the load online to adaptively evaluate this function.

\subsection{Control Formulation}
At control step $t$, DANC takes the measured elongation history and the desired-trajectory preview as its inputs, and outputs the pressure command sent to the proportional valve. The load variable is expressed in mass-equivalent units, $m_t=T_t/g$, where $T_t$ is the tensile load and $g$ is the gravitational acceleration. DANC's neural maps are trained to predict this real physical load directly; at run time, the online estimate $\hat m_t$ then acts as an internal adaptive coordinate rather than a calibrated measurement. The main notation used in this section is summarized in Table~\ref{tab:danc-notation}.

\begin{table}[!t]
    \caption{Notation Used in the DANC Formulation}
    \label{tab:danc-notation}
    \centering
    \setlength{\tabcolsep}{0.06in}
    \renewcommand{\arraystretch}{1.06}
    \begin{tabular}{>{\centering\arraybackslash}m{0.25\columnwidth}m{0.58\columnwidth}}
    \toprule
        Symbol & Meaning \\
    \midrule
        $x_t$, $x_t^{*}$ & measured and desired elongation \\
        $p_t$ & command pressure \\
        $m_t$, $\hat m_t$ & load state and its estimate \\
        $\mathbf{x}_{t}^{\mathrm{his}}$, $\mathbf{x}_{t}^{\mathrm{pre}}$ & history and preview elongation windows \\
        $n,h,k,l$ & temporal-window length parameters \\
        $\mathcal{M}$, $N$ & predefined load set and number of loads \\
        $\mathcal{P}_i$, $\mathcal{L}$, $\hat p(t,m)$ & neural maps and pressure--load profile \\
        $\tilde m_t$, $m_t^{\mathrm{d}}$, $m_t^{\mathrm{s}}$ & load quantities: raw MLP estimate; filtered dynamic estimate; static compensation state \\
    \bottomrule
    \end{tabular}
\end{table}

In a static and history-free setting, any two of $(p,x,m)$ determine the third. In practice, however, the pressure required for a desired elongation is affected by nonideal actuator dynamics, including quasi-static and rate-dependent hysteresis, creep, and pneumatic delay. DANC therefore represents PAM control by a load-conditioned inverse map:
\begin{equation}
    \hat p(t,m) =
    f\left(\mathbf{x}_{t}^{\mathrm{his}},
    \mathbf{x}_{t}^{\mathrm{pre}};
    m\right),
\end{equation}
where
\begin{equation}
\begin{aligned}
    \mathbf{x}_{t}^{\mathrm{pre}}
    &=(x_t^{*},x_{t+1}^{*},\ldots,x_{t+h}^{*}),\\
    \mathbf{x}_{t}^{\mathrm{his}}
    &=(x_{t-n},x_{t-n+1},\ldots,x_{t-1}).
\end{aligned}
\end{equation}
The measured-history window captures actuator memory associated with quasi-static and rate-dependent hysteresis and creep. The preview window provides the upcoming motion trend, allowing pressure changes to be issued ahead of the delayed pneumatic response. Since the load state is usually not measured on the arm, PIC constructs the pressure--load function $\hat p(t,m)$ and LEC supplies the load estimate at which it is evaluated.

\subsection{Pressure Inference Channel}
For a given motion state, the PIC estimates how the required pressure changes with load. It is trained from data collected under a finite set of predefined loads:
\begin{equation}
\begin{aligned}
    \mathcal{M}&=\{m_1,m_2,\ldots,m_N\},\\
    m_1&\leq m_{\min}<m_2<\cdots<m_{N-1}<m_{\max}\leq m_N .
\end{aligned}
\end{equation}
where $[m_{\min},m_{\max}]$ denotes the expected operating load range.
For each load $m_i$, an MLP $\mathcal{P}_i$ is trained from dynamic pressure--elongation data collected under that fixed load; the data-collection procedure is detailed in Section~\ref{subsec:danc-training}. Given the temporal elongation input, the network predicts the pressure that would be required if the PAM were operating under $m_i$:
\begin{equation}
    \hat p_i(t)=\mathcal{P}_i
    \left(\mathbf{x}_{t}^{\mathrm{his}},
    \mathbf{x}_{t}^{\mathrm{pre}}\right),
    \quad i=1,\ldots,N .
\end{equation}
The $N$ outputs are then regarded as samples of the pressure required at the same motion state but under different loads. A polynomial is fitted along the load coordinate,
\begin{equation}
    \hat p(t,m)=\sum_{r=0}^{M} \beta_r(t)m^r ,
    \qquad M\leq N-1 .
\end{equation}
In our implementation, we use the predefined load set $\{0,2,4,6,8\}$ kg ($N=5$) and set the polynomial degree to $M=4$. This construction keeps load as an explicit operating coordinate: the load-specific MLPs predict the required pressures, and the polynomial extends them to $\hat p(t,m)$ at any online load estimate.

Once the load estimate $\hat m_t$ is available, the command sent to the proportional valve is
\begin{equation}
    p_t =
    \mathrm{sat}_{[0,p_{\max}]}
    \left(\hat p(t,\hat m_t)\right).
\end{equation}

\subsection{Load Estimation Channel}
The LEC provides the load coordinate $\hat m_t$ for evaluating the PIC pressure--load function. As a control state, $\hat m_t$ follows payload changes and disturbances to improve elongation tracking rather than report a calibrated external load. The MLP $\mathcal{L}$ supplies the primary estimate from dynamic data, and a static compensator removes residual steady-state elongation error.

The MLP estimates load from recent measured elongation and pressure histories available at the beginning of the control step:
\begin{equation}
    \tilde m_t =
    \mathcal{L}\left(
    x_{t-k},\ldots,x_{t-1},
    p_{t-l},\ldots,p_{t-1}
    \right).
\end{equation}
Because $\mathcal{L}$ is trained on dynamic fixed-load sequences, $\tilde m_t$ responds to sudden payload changes and external forces. The raw estimate is smoothed by a scalar Kalman filter:
\begin{equation}
    \begin{aligned}
    m_{t|t-1}^{\mathrm{d}}
    &=m_{t-1}^{\mathrm{d}},\\
    m_t^{\mathrm{d}}
    &=m_{t|t-1}^{\mathrm{d}}
    +K_t\left(\tilde m_t-m_{t|t-1}^{\mathrm{d}}\right),
    \end{aligned}
\end{equation}
where $K_t$ is the Kalman gain under a random-walk load model. This dynamic estimate provides fast adaptation, but estimation bias, material drift, or specimen variation can still leave a nonzero elongation error when the load is constant or slowly varying.

The static compensator converts this residual error into a correction of the load coordinate. Using the pressure-dependent static PAM model in Section~\ref{sec:characterization}, specifically the effective-force coordinate in \eqref{eq:pressure-dependent-effective-force} and the fitted static relation in \eqref{eq:static-pam-model}, the compensator first inverts the polynomial part of the model. For a given elongation $x$, $F_{\mathrm{eff}}$ is obtained from
\begin{equation}
    \sum_{j=0}^{5}a_j F_{\mathrm{eff}}^{j}-x=0 .
\end{equation}
Among the real roots, the physically admissible root in the operating range is selected. This defines the inverse static map $\mathcal{F}(x)$ from elongation to effective-force coordinate. Under pressure $p$, the corresponding tensile load is
\begin{equation}
    T(x,p)=
    \max\left(0,\,
    \mathcal{F}(x)-S_0p-\alpha_p p^2
    \right),
\end{equation}
We denote the corresponding static load map by $\Phi(x,p)$:
\begin{equation}
    \Phi(x,p)=\frac{T(x,p)}{g}.
\end{equation}
The static discrepancy compares the loads inferred from the measured and desired elongations under the current commanded pressure:
\begin{equation}
    e_t^{\mathrm{s}}
    =
    \Phi(x_{t-1},p_{t-1})-\Phi(x_{t-1}^{*},p_{t-1}).
\end{equation}
This discrepancy drives the static compensation state through a PI recursion:
\begin{equation}
    I_t^{\mathrm{s}}=
    I_{t-1}^{\mathrm{s}}+e_t^{\mathrm{s}},
\end{equation}
\begin{equation}
    m_t^{\mathrm{s}} =
    K_p e_t^{\mathrm{s}}
    +K_i I_t^{\mathrm{s}} .
\end{equation}
Since the required pressure decreases with the load coordinate, a measured elongation below the target gives $e_t^{\mathrm{s}}<0$, which lowers $m_t^{\mathrm{s}}$ and raises the commanded pressure, so the loop corrects in the right direction.
The load coordinate used by the PIC is obtained by combining the dynamic estimate and static compensation state:
\begin{equation}
    \hat m_t =
    \alpha m_t^{\mathrm{d}}
    +(1-\alpha)m_t^{\mathrm{s}},
\end{equation}
where $0\leq\alpha\leq1$. The resulting $\hat m_t$ should be interpreted as a latent adaptation coordinate. If aging, friction change, or replacement of a PAM changes the actuator constitutive relation, the unobserved mismatch can be absorbed as a shift in $\hat m_t$. The controller can therefore sacrifice physical load accuracy to choose a pressure that better tracks $x_t^{*}$, which is the main source of adaptivity in the DANC.

\subsection{Training and Deployment}
\label{subsec:danc-training}
The data-collection procedure used to train DANC is summarized in Algorithm~\ref{alg:danc}. Offline training uses two types of data. First, dynamic pressure--elongation sequences are collected at 25 Hz under the fixed loads in $\mathcal{M}$. For each load, the pressure command is randomized over 0--450 kPa using alternating holding intervals and smooth sinusoidal transitions, so that the dataset includes holding, slow motion, and rapid inflation/deflation. In our implementation, the pressure-change rate is sampled from 40--600 kPa/s, and approximately 20 minutes of data are collected for each load. Static data are collected separately from 0--450 kPa pressure steps after the transient response has settled. Because the arm uses both standard and short PAMs, data collection, MLP training, and compensator fitting are performed separately for the two PAM types, while the controller workflow is unchanged.

\begin{algorithm}[!t]
\caption{Data Collection for Training DANC}
\label{alg:danc}
\begin{algorithmic}[1]
\Require Load set $\mathcal{M}$, pressure range $[0,p_{\max}]$, rate range $[v_{\min},v_{\max}]$
\Ensure Dynamic dataset $\mathcal{D}_{\mathrm{dyn}}$, static dataset $\mathcal{D}_{\mathrm{stat}}$
\For{each load $m_i \in \mathcal{M}$}
    \While{dynamic collection time is not reached}
        \State Sample target pressure $p^{\mathrm{tar}}\in[0,p_{\max}]$ and rate $v\in[v_{\min},v_{\max}]$.
        \State Move from current pressure to $p^{\mathrm{tar}}$ with a smooth sinusoidal transition, then hold.
        \State Record $(p_t,x_t,m_i)$ into $\mathcal{D}_{\mathrm{dyn}}$.
    \EndWhile
    \State Apply pressure steps; after each step settles, record steady $(p,x,m_i)$ into $\mathcal{D}_{\mathrm{stat}}$.
\EndFor
\end{algorithmic}
\end{algorithm}

The fixed-load dynamic sequences train both neural components. Each PIC network $\mathcal{P}_i$ uses three hidden layers of 100 ReLU units and predicts $p_t$ from $\mathbf{x}_{t}^{\mathrm{his}}$ and $\mathbf{x}_{t}^{\mathrm{pre}}$. During training, the preview is formed from recorded future elongations; during deployment, it is replaced by the planned desired-trajectory preview. The LEC network $\mathcal{L}$ uses four hidden layers of 100 ReLU units and is trained on the same sequences with the fixed load as the target. The temporal-window parameters are selected by training candidate combinations and choosing, for each network, the combination that minimizes validation-set RMSE, yielding $n=20$, $h=50$, $k=19$, and $l=9$.

The static data identify the model used by $\Phi(x,p)$ and support rapid static calibration for different PAM specimens. The Kalman noise parameters and the gains $K_p$, $K_i$, and $\alpha$ are selected in closed-loop load-change experiments; here, $K_p=0.2$, $K_i=0.005$, and $\alpha=0.4$. During online deployment, joint- or arm-level trajectories are mapped to desired PAM elongations, and each PAM bundle runs DANC to command $p_t=\mathrm{sat}_{[0,p_{\max}]}(\hat p(t,\hat m_t))$.

\begin{figure*}[!t]
\centering
\includegraphics[width=0.98\textwidth]{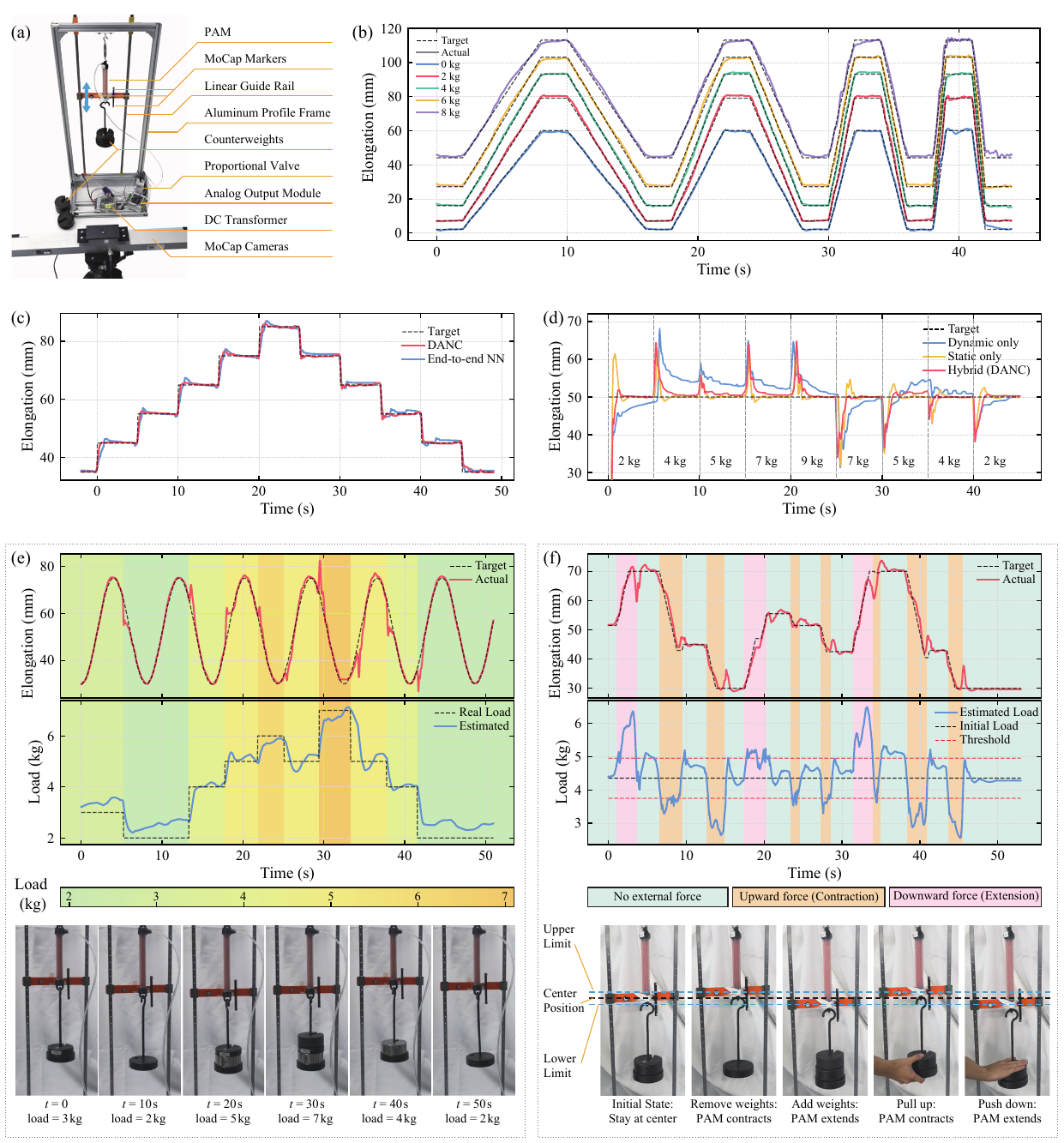}
\caption{Single-PAM load-adaptive tracking and controller ablations. (a) Experimental setup. (b) Ramp tracking under fixed loads from 0 to 8 kg. (c) Tracking of consecutive 10-mm steps under a 5-kg load by DANC and an end-to-end neural-network ablation. (d) LEC ablation at a 50-mm target under the load sequence 2, 4, 5, 7, 9, 7, 5, 4, and 2 kg. (e) Sinusoidal tracking under time-varying loads, with real and estimated load curves. (f) Responses to load changes and manual upward and downward forces.}
\label{fig:muscle-control-main}
\end{figure*}

\section{Experiments}
\label{sec:experiments}
The experiments proceed from a single PAM to a spherical joint and the full arm. The actuator-level tests evaluate tracking under fixed and changing loads, controller ablations, cross-specimen calibration, and long-term drift. The joint- and arm-level tests evaluate trajectory tracking, motion range, variable stiffness, interaction, coordinated 7-DOF motion, and operation after one PAM is removed.

\subsection{Single-PAM Load-Adaptive Tracking}
The first set of experiments evaluates DANC on a single PAM. As shown in Fig.~\ref{fig:muscle-control-main}(a), the PAM is mounted vertically on a linear guide, which suppresses pendulum-like swinging of the hanging load. External weights provide the axial load, an OptiTrack V120 Trio motion-capture system measures elongation, and a Festo VPPE proportional valve supplies the commanded pressure. This setup isolates the actuator-level problem: for a desired elongation trajectory, the controller must choose pressure while the tensile load changes.

Fig.~\ref{fig:muscle-control-main}(b) evaluates ramp tracking under fixed loads of 0, 2, 4, 6, and 8 kg. The target range for each load is selected from the feasible elongation region in Figs.~\ref{fig:character-muscle}(d) and (e), and the successive ramps increase in speed. DANC closely follows the targets across all tested loads and speeds.

Fig.~\ref{fig:muscle-control-main}(c) compares DANC with an end-to-end neural-network ablation under a 5-kg load. The target elongation starts at 35~mm, increases to 85~mm in 10~mm increments, and then decreases to 35~mm in the same increments; each transition is applied as a step command. The ablation replaces the PIC, LEC, and their coupling with a single MLP. The MLP directly receives the same vector available to DANC's computational modules after the temporal extensions $(x_{t-n},\ldots,x_{t-1},x^*_{t},\ldots,x^*_{t+h},p_{t-l},\ldots,p_{t-1})$ as input and outputs the next-step pressure command $p_{t}$ as does DANC. Averaged over all step transitions, DANC achieves a 0.17-s rise time, a 0.62-s settling time under the 5\% criterion, 7.9\% overshoot, and a 0.13-mm steady-state error. The end-to-end network is slower, fails to satisfy the 5\% settling criterion, and produces larger overshoot and steady-state error.

We compare DANC with prior studies that (i) tested single-PAM step responses experimentally and (ii) reported the corresponding conditions and results (Table~\ref{tab:comparison}). However, direct comparison across PAM controllers is confounded by differences in load and PAM's mechanical properties such as mass, stiffness and damping. We therefore scale each measured response against a local second-order reference,
\begin{equation}
    m_{\mathrm{eff}}\ddot{y}_{\mathrm{SO}}
    +2\zeta_{\mathrm{eq}}
    \sqrt{m_{\mathrm{eff}}k_{\mathrm{eff}}}\dot{y}_{\mathrm{SO}}
    +k_{\mathrm{eff}}y_{\mathrm{SO}}
    =k_{\mathrm{eff}}r,
    \label{eq:comparison-second-order}
\end{equation}
where $r$ and $y_{\mathrm{SO}}$ are the normalized elongation command and response, respectively, $m_{\mathrm{eff}}$ is the moving mass, and $k_{\mathrm{eff}}=\Delta F/\Delta x$ is the mean axial stiffness over the step range.

Here $m_e$ is the external load and $m_i$ is the PAM mass. Because $m_i$ is concentrated near its two end fittings, we assign half of it to the moving end:
\begin{equation}
    m_{\mathrm{eff}}=m_e+\frac{1}{2}m_i.
    \label{eq:comparison-effective-mass}
\end{equation}
This approximation is applied consistently and excludes unreported sliders, connectors, and sensors.

The effective mass $m_{\mathrm{eff}}$ and mean stiffness $k_{\mathrm{eff}}$ can be estimated from the reported test conditions, whereas the damping ratio generally cannot. For an underdamped second-order system, the overshoot $M_p$ depends only on the damping ratio. We therefore infer an equivalent damping ratio $\zeta_{\mathrm{eq}}$ from $M_p$, rather than compare the overshoot directly across studies:
\begin{equation}
\begin{aligned}
    \zeta_{\mathrm{eq}}
    &=\frac{-\ln M_p}
      {\sqrt{\pi^2+\left(\ln M_p\right)^2}},\\
    \omega_n&=\sqrt{\frac{k_{\mathrm{eff}}}{m_{\mathrm{eff}}}},
\end{aligned}
\label{eq:comparison-damping}
\end{equation}
The corresponding underdamped unit-step response is
\begin{equation}
\begin{aligned}
    y_{\mathrm{SO}}(t)
    &=1-e^{-\zeta_{\mathrm{eq}}\omega_n t}
    \left[
    \cos(\omega_dt)\right.\\
    &\qquad\left.
    +\frac{\zeta_{\mathrm{eq}}}
    {\sqrt{1-\zeta_{\mathrm{eq}}^2}}
    \sin(\omega_dt)
    \right],\\
    \omega_d&=\omega_n\sqrt{1-\zeta_{\mathrm{eq}}^2}.
\end{aligned}
\label{eq:comparison-step-response}
\end{equation}
Using the same 10\%--90\% rise-time and 5\% settling-time definitions as the measured data, the reference times are
\begin{equation}
\begin{aligned}
    t_{r,\mathrm{SO}}&=t_{90}-t_{10},\\
    y_{\mathrm{SO}}(t_{10})&=0.1,\quad
    y_{\mathrm{SO}}(t_{90})=0.9,\\
    t_{s,\mathrm{SO}}
    &=\inf\left\{t:\,
    |1-y_{\mathrm{SO}}(\tau)|\leq0.05,\ 
    \forall\tau\geq t\right\}.
\end{aligned}
\label{eq:comparison-reference-times}
\end{equation}
These times are evaluated numerically. The reported dimensionless ratios are
\begin{equation}
    \rho_r=\frac{t_r}{t_{r,\mathrm{SO}}},
    \qquad
    \rho_s=\frac{t_s}{t_{s,\mathrm{SO}}}.
    \label{eq:comparison-normalized-times}
\end{equation}
These ratios compare the measured closed-loop rise and settling times with those of the corresponding equivalent second-order system; lower values therefore indicate better relative control performance.
This reference serves only as a common scale for cross-study comparison, not as a complete pneumatic-actuator model.
\begin{table*}[!t]
    \caption{Plant-Normalized Experimental Step-Response Comparison of Single-PAM Position Controllers}
    \label{tab:comparison}
    \centering
    \scriptsize
    \setlength{\tabcolsep}{0.04in}
    \renewcommand{\arraystretch}{1.10}
    \newcommand{\comparisonhead}[2]{\shortstack[c]{\strut #1\\\strut #2}}
    \begin{tabular}{@{}llccccccccc@{}}
    \toprule
        & & \multicolumn{5}{c}{Test conditions} & \multicolumn{2}{c}{Damping characteristics} & \multicolumn{2}{c}{Normalized time ratios} \\
    \cmidrule(lr){3-7}\cmidrule(lr){8-9}\cmidrule(l){10-11}
        \comparisonhead{Method}{} & \comparisonhead{PAM}{} &
        \comparisonhead{External load}{$m_e$ (kg)} &
        \comparisonhead{PAM mass}{$m_i$ (kg)} &
        \comparisonhead{Effective mass}{$m_{\mathrm{eff}}$ (kg)} &
        \comparisonhead{Mean stiffness}{$k_{\mathrm{eff}}$ (N/mm)} &
        \comparisonhead{Step size}{$\Delta x$ (mm)} &

        \comparisonhead{Overshoot}{$M_p$ (\%)} &
        \comparisonhead{Equivalent}{damping $\zeta_{\mathrm{eq}}$} &

        \comparisonhead{Rise-time}{ratio $\rho_r$} &
        \comparisonhead{Settling-time}{ratio $\rho_s$} \\
    \midrule
        PI + HC\cite{minh2010cascade} & Festo MAS-20 & 6 & 0.40 & 6.20 & 16.99 & 10 & 5.7 & 0.67 & 7.67 & 13.04 \\ 
        CFTOC\cite{andrikopoulos2013piecewise} & Festo DMSP-40 & 0 & 0.84 & 0.42 & 12.48 & 20 & 3.4 & 0.73 & 31.70 & 28.89 \\ 
        Integral control\cite{tondu2014robust} & Custom McKibben & 5 & NR & $<5.50$ & 8.33 & 30 & 3.2 & 0.74 & $>9.86$ & $>11.41$ \\ 
        PID + AMC\cite{qin2022active} & Festo DMSP-20 & 0 & 0.22 & 0.11 & 18.97 & 10 & 2.5 & 0.76 & 32.13 & 36.56 \\ 
        IDA-PBC\cite{tanaka2025position} & Custom McKibben & 1.25 & NR & $<1.75$ & 0.86 & 20 & 20.8 & 0.45 & $>1.58$ & $>2.32$ \\ 
        \textbf{DANC (ours)} & Custom extensile PAM & 5 & 0.07 & 5.03 & 0.74 & 10 & 7.9 & 0.63 & \textbf{1.07} & \textbf{1.46} \\
        \shortstack[l]{End-to-end NN (ablation)} & Custom extensile PAM & 5 & 0.07 & 5.03 & 0.74 & 10 & 22.1 & 0.43 & 2.96 & -- \\
    \bottomrule
    \end{tabular}
    \vspace{2pt}

    \begin{minipage}{0.99\textwidth}
        \footnotesize\textit{Note:} Prior-controller data were digitized from the cited experimental plots. For DANC and the end-to-end NN, the measured metrics and stepwise stiffnesses are first averaged over all 10-mm steps in Fig.~\ref{fig:muscle-control-main}(c); the derived quantities are then calculated. When PAM mass is unreported, $m_i<1$ kg is used as a conservative upper bound, consistent with the sub-1-kg masses reported for the other PAMs. NR denotes ``not reported,'' and ``--'' denotes an unavailable metric.
    \end{minipage}
\end{table*}

\begin{table}[!t]
    \caption{Ablation of the Load Estimation Channel under Stepwise Load Changes}
    \label{tab:lec-ablation}
    \centering
    \scriptsize
    \setlength{\tabcolsep}{0.06in}
    \renewcommand{\arraystretch}{1.12}
    \begin{tabular}{@{}lcccc@{}}
    \toprule
        Method &
        \shortstack{Overshoot\\(mm)} &
        \shortstack{RMSE\\(mm)} &
        \shortstack{Mean recovery\\time $\pm1$ mm (s)} &
        \shortstack{Mean recovery\\time $\pm2$ mm (s)} \\
    \midrule
        Dynamic only & 2.39 & 4.81 & 3.84 & 2.12 \\
        Static only & 2.02 & 2.88 & \textbf{1.36} & 1.06 \\
        \textbf{Hybrid (DANC)} & \textbf{0.69} & \textbf{2.83} & 1.66 & \textbf{0.93} \\
    \bottomrule
    \end{tabular}
    \vspace{2pt}

    \begin{minipage}{0.45\textwidth}
        \footnotesize\textit{Note:} The target elongation is 50 mm, and the load sequence is 2, 4, 5, 7, 9, 7, 5, 4, and 2 kg. RMSE is computed over the full trial; overshoot and recovery time are averaged over the eight load transitions in Fig.~\ref{fig:muscle-control-main}(d).
    \end{minipage}
\end{table}

Table~\ref{tab:comparison} shows that DANC has the smallest time ratios among the controllers selected for comparison: $\rho_r=1.07$ and $\rho_s=1.46$. The end-to-end ablation increases $\rho_r$ to 2.96, raises overshoot from 7.9\% to 22.1\%, and fails to satisfy the 5\% settling criterion, supporting the dual-channel design.

Fig.~\ref{fig:muscle-control-main}(d) and Table~\ref{tab:lec-ablation} isolate the two LEC components while the target remains at 50 mm and the load follows 2, 4, 5, 7, 9, 7, 5, 4, and 2 kg. ``Dynamic only'' retains the MLP estimate and Kalman filter, ``Static only'' retains the static compensator, and Hybrid uses both. Hybrid gives the lowest overshoot and RMSE and the shortest $\pm2$-mm recovery time; Static only is slightly faster within the tighter $\pm1$-mm band. The two components therefore provide complementary transient and steady-state correction.

Fig.~\ref{fig:muscle-control-main}(e) tests sinusoidal tracking under time-varying load. The desired elongation is
\begin{equation}
    x^*(t)=52.5+22.5\sin\left(\frac{2\pi}{8}t\right)\ \mathrm{mm},
\end{equation}
with a range selected from Figs.~\ref{fig:character-muscle}(d) and (e). The elongation remains close to the target as weights are added and removed, while the LEC follows the load trend. Short spikes occur at the discontinuous load changes.

Fig.~\ref{fig:muscle-control-main}(f) applies interaction-like disturbances about an initial 4-kg load. A $\pm0.6$-kg tolerance band triggers motion only when the estimate exceeds 5 kg or falls below 3 kg. Removing weight or pulling upward triggers contraction; adding weight or pushing downward triggers extension. The estimate changes consistently with these actions, indicating that the LEC can provide an interaction cue without a force sensor.

\subsection{Single-PAM Calibration}
\begin{figure*}[!t]
\centering
\includegraphics[width=\textwidth]{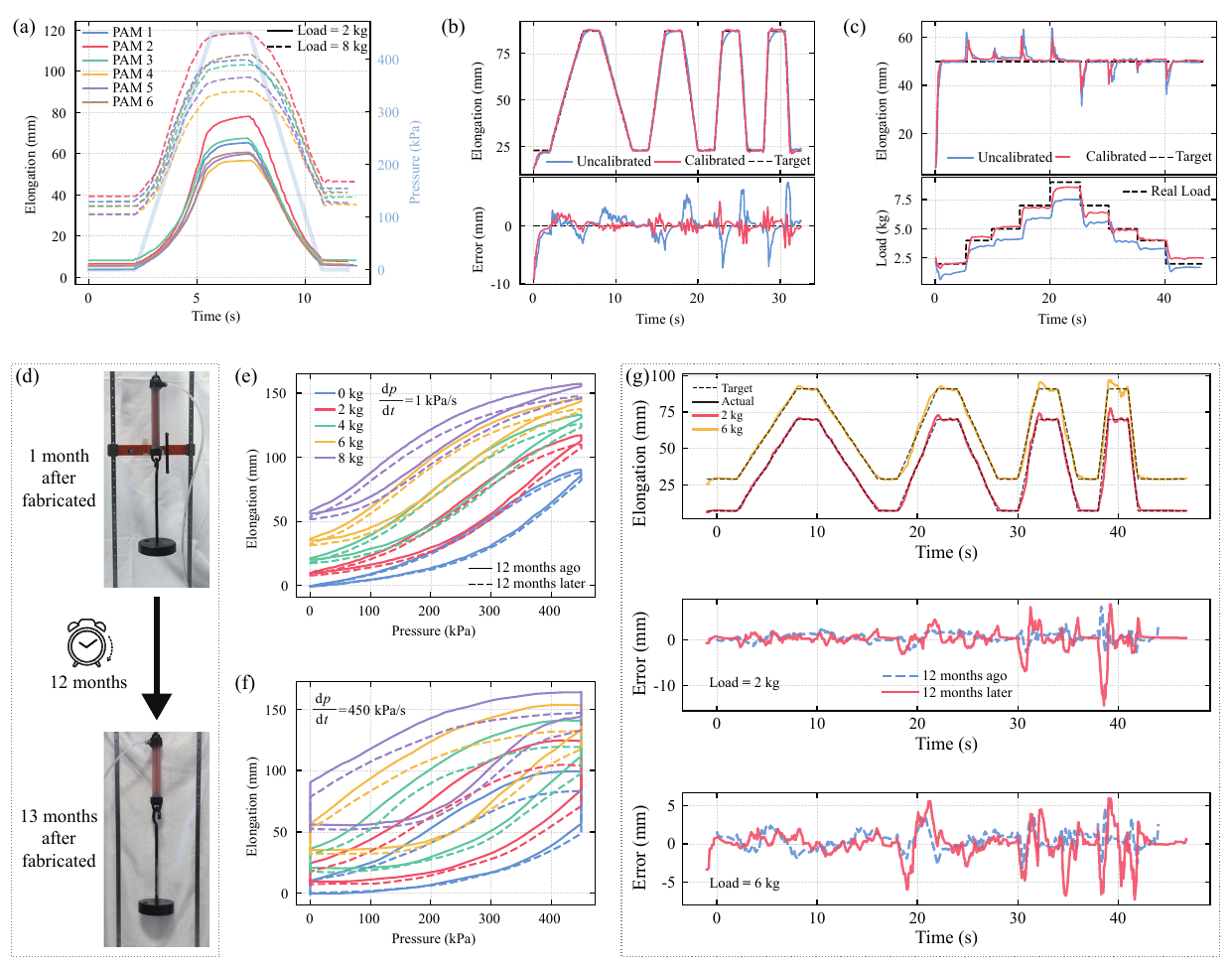}
\caption{Single-PAM specimen variation, calibration, and long-term drift. (a) Responses of six same-batch PAMs to a common pressure input under 2- and 8-kg loads. PAM 3 is the training reference, and PAM 2 is selected for calibration because it shows the most pronounced mismatch with PAM 3. (b) Ramp tracking under a 5-kg load before and after calibration. (c) Elongation regulation and load estimation before and after calibration under the load sequence 2, 4, 5, 7, 9, 7, 5, 4, and 2 kg. (d) Aging-test setup at one and 13 months after fabrication. (e) Quasi-static pressure--elongation hysteresis before and after the 12-month interval. (f) Rate-dependent pressure--elongation hysteresis at 450 kPa/s before and after the interval. (g) Tracking trajectories and errors under 2- and 6-kg loads before and after the interval.}
\label{fig:muscle-control-variation}
\end{figure*}

Fig.~\ref{fig:muscle-control-variation}(a) applies the same pressure input to six same-batch PAMs under 2- and 8-kg loads. Their different steady elongations and dynamic amplitudes motivate a rapid calibration intended to reduce the need for actuator-specific retraining.

The calibration uses a stiffness-ratio correction based on a single static pressure test. Let $\Delta x_0$ be the static elongation of the reference PAM used for training at 450 kPa under an 8 kg load, and let $\Delta x_c$ be the corresponding elongation of the PAM to be calibrated under the same condition. The calibration coefficient is
\begin{equation}
    \beta=\frac{\Delta x_0}{\Delta x_c}.
\end{equation}
All measured and desired elongation signals entering both PIC and LEC are scaled as
\begin{equation}
    \tilde{x}_t=\beta x_t,\qquad
    \tilde{x}_t^{*}=\beta x_t^{*},
\end{equation}
while the pressure signals and trained parameters remain unchanged. This procedure maps the target and measured elongations of the calibrated PAM into the elongation scale of the reference PAM, thereby adjusting the constitutive relation seen by the controller using only a short static measurement.


With PAM 3 as the training reference, PAM 2---the specimen showing the most pronounced difference in the plotted responses in Fig.~\ref{fig:muscle-control-variation}(a)---was selected for the calibration tests in Fig.~\ref{fig:muscle-control-variation}(b) and (c), 
evaluating the calibration method under a fixed 5-kg load; calibration reduces the ramp-tracking error, particularly near rapid reversals. Under the same fixed-target, stepwise-load protocol as Fig.~\ref{fig:muscle-control-main}(d), Fig.~\ref{fig:muscle-control-variation}(c) shows that calibration reduces the elongation transients and brings the estimated load closer to the applied load.

\subsection{Long-Term Drift}
The same PAM was tested one and 13 months after fabrication using the setup in Fig.~\ref{fig:muscle-control-variation}(d). Both the quasi-static curves in Fig.~\ref{fig:muscle-control-variation}(e) and the 450-kPa/s loops in Fig.~\ref{fig:muscle-control-variation}(f) change after 12 months, showing drift in the pressure--elongation relation.

Fig.~\ref{fig:muscle-control-variation}(g) compares tracking under 2- and 6-kg loads before and after the interval, without recalibration. Despite the constitutive drift, the post-aging responses remain close to the original results, indicating that DANC tolerates the residual change.

\begin{figure*}[!t]
\centering
\includegraphics[width=\textwidth]{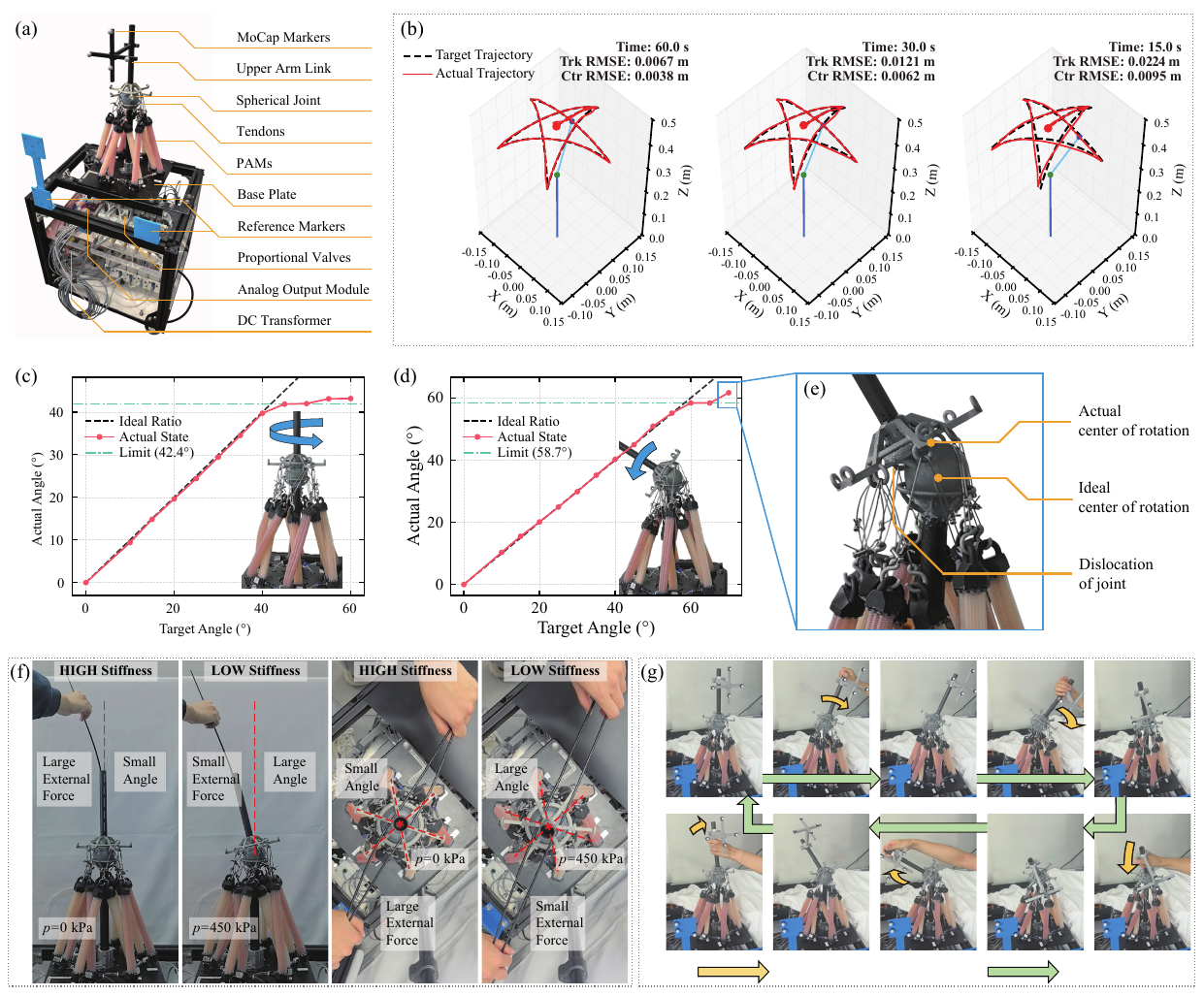}
\caption{Single-joint trajectory tracking, variable stiffness, and drag-following interaction. (a) Experimental setup for single-joint trajectory tracking. (b) Three-dimensional star-path tracking at three different speeds. (c) Torsional-DOF tracking and measured rotation limit. (d) Bending-DOF tracking and measured rotation limit. (e) Recoverable large-angle joint dislocation between the ideal and actual centers of rotation. (f) Variable-stiffness capability under external forces for low- and high-pressure states. (g) Drag-following interaction; yellow arrows indicate the applied external force, and green arrows indicate the time sequence of the compliant motion.}
\label{fig:joint-control}
\end{figure*}

\subsection{Spherical-Joint Trajectory Tracking}
The spherical-joint experiments test whether actuator-level control can coordinate antagonistic PAM groups through a redundant joint. The joint module is fixed to the world frame $\{W\}$, the upper-arm link is attached to the moving joint cap, and motion-capture markers measure the link pose, as shown in Fig.~\ref{fig:joint-control}(a). For a desired joint angle, the joint kinematics and PAM routing geometry in Section~\ref{sec:characterization} are first used to compute the target elongation of each PAM. Each PAM then runs DANC independently to track its assigned elongation.

To express the measured motion in a local coordinate system, the world frame $\{W\}$ is constructed from the three fixed reference markers on the blue plates in Fig.~\ref{fig:joint-control}(a). The three markers lie on one plane, which defines the $Y_W$--$Z_W$ plane of $\{W\}$, with $P_1$ as the origin and $P_2-P_1$ defining the $Y_W$ direction. Let the marker positions in the camera frame $\{C\}$ be $P_1$, $P_2$, and $P_3$. Then
\begin{equation}
    \mathbf{y}_W = \frac{P_2-P_1}{\|P_2-P_1\|}, \quad
    \mathbf{x}_W =
    \frac{\mathbf{y}_W\times(P_3-P_1)}
    {\|\mathbf{y}_W\times(P_3-P_1)\|},
\end{equation}
and $\mathbf{z}_W=\mathbf{x}_W\times\mathbf{y}_W$. All markers are defined as rigid bodies, and their positions are taken as the corresponding rigid-body centroids. If $\mathbf{p}_{b}^{C}$ and $\mathbf{q}_{b}^{C}$ are the centroid position and orientation quaternion of the moving rigid body in the camera frame, its pose in the world frame is
\begin{equation}
    \mathbf{p}_{b}^{W}=R_{CW}^{T}(\mathbf{p}_{b}^{C}-P_1),
    \quad
    \mathbf{q}_{b}^{W}=\mathbf{q}_{WC}\otimes\mathbf{q}_{b}^{C},
\end{equation}
where $R_{CW}=[\mathbf{x}_W,\mathbf{y}_W,\mathbf{z}_W]$, $\mathbf{q}_{WC}$ is the quaternion corresponding to $R_{CW}^{T}$
. This gives joint-space tracking measurements independent of camera-frame placement.

Fig.~\ref{fig:joint-control}(b) shows three-dimensional star-path tracking at three speeds. The trajectory RMSEs are 6.7, 12.1, and 22.4 mm, and the contour RMSEs are 3.8, 6.2, and 9.5 mm, where the contour error is computed from the nearest point on the other trajectory. Error increases as the motion becomes faster, which is consistent with pneumatic delay and rate-dependent PAM behavior; in the joint test, multi-PAM coordination and spherical-joint friction further contribute to the deviation. The measured paths still preserve the star shape, showing that the actuator-level inverse controller can be composed across several PAMs to produce coordinated joint motion.

Figs.~\ref{fig:joint-control}(c) and (d) evaluate the available torsion and bending ranges by commanding increasing target angles. In the central range, the measured angle follows the target approximately linearly. As the command approaches the limit, the joint reaches a static equilibrium at the maximum available pressure, so increasing the target no longer produces the same increase in measured angle. The measured limits are about $42.4^\circ$ for torsion and $58.7^\circ$ for bending. These values can be compared with the static prediction in Fig.~\ref{fig:character-arm}(a); the measured ranges are smaller because the model does not include all resistance from the silicone membrane and friction between the joint cup and head. At still larger bending angles, Fig.~\ref{fig:joint-control}(e) shows the actual center of rotation separating from the ideal spherical center. Rather than normal tracking, this large-angle dislocation acts as a self-protection mechanism: the recoverable displacement releases the geometric constraint and avoids excessive local stress in the membrane and cup--head contact region.

Fig.~\ref{fig:joint-control}(f) demonstrates the variable-stiffness capability of the joint. External forces are manually applied to the link under two pressure states. At 0 kPa, the PAMs carry higher tensile force, and even a relatively large external force produces only a small joint deflection, indicating high overall stiffness. At 450 kPa, the PAM tension is lower, and a smaller external force can produce a larger deflection, indicating a more compliant joint. The same trend is observed in both bending and torsion.

\begin{figure*}[!t]
\centering
\includegraphics[width=\textwidth]{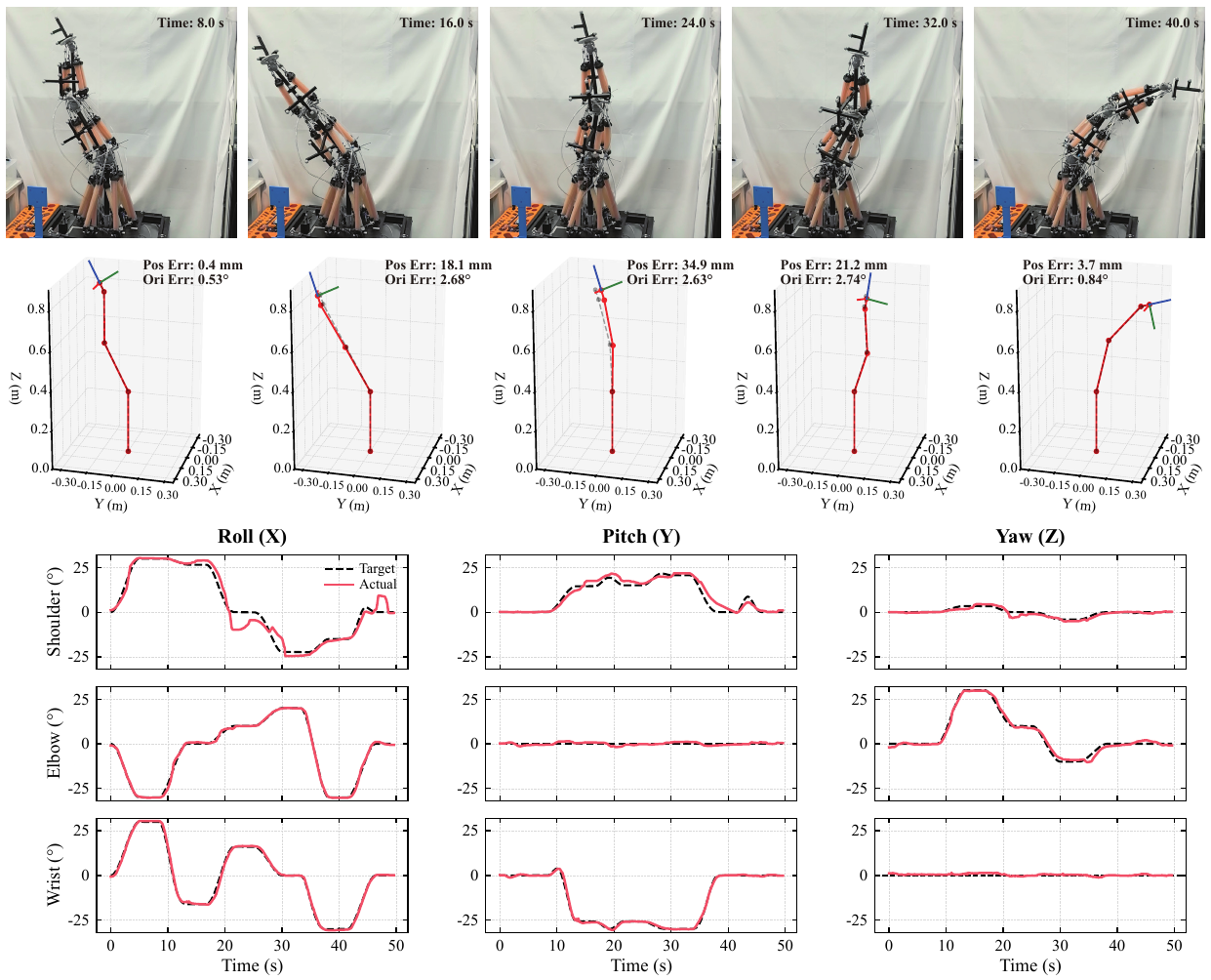}
\caption{Full-arm coordinated motion control and end-effector accuracy. The top two rows pair physical snapshots with reconstructed poses and the corresponding end-effector errors for representative arm postures. The bottom plots compare target and measured joint angles for shoulder, elbow, and wrist rotations about the roll (X), pitch (Y), and yaw (Z) axes.}
\label{fig:arm-control-main}
\end{figure*}

\subsection{Drag-Following Interaction}
Fig.~\ref{fig:joint-control}(g) shows a drag-following experiment based on the PAM load estimates. No external force sensor is used. At each control step, the LEC provides a mass-equivalent load estimate $\hat m_i$ for each PAM. This estimate is converted to the estimated tendon tension $\hat T_i=g\hat m_i$. The estimated PAM force applied to the moving link is
\begin{equation}
    \hat{\mathbf{f}}_i
    =\hat T_i\frac{\mathbf{b}_i-\mathbf{a}_i(\mathbf{q})}
    {\ell_i(\mathbf{q})},
\end{equation}
and the corresponding estimated PAM torque is
\begin{equation}
    \hat{\boldsymbol{\tau}}_{\mathrm{PAM}}(\mathbf{q})
    =
    \sum_{i=1}^{n_{\mathrm{PAM}}}\mathbf{a}_i(\mathbf{q})\times\hat{\mathbf{f}}_i .
\end{equation}
For a simple interaction cue, gravitational and passive joint torques are neglected. Under this quasi-static approximation, the external torque is estimated as
\begin{equation}
    \hat{\boldsymbol{\tau}}_{\mathrm{ext}}
    =
    -\hat{\boldsymbol{\tau}}_{\mathrm{PAM}}(\mathbf{q}).
\end{equation}
The resulting quantity is used only to detect the direction and onset of interaction, rather than as a calibrated measurement of external torque.
When the magnitude of the relevant component of $\hat{\boldsymbol{\tau}}_{\mathrm{ext}}$ exceeds a preset threshold, the target joint angle is updated in the corresponding direction. In the experiment, the triggered motion is limited to bending-angle changes or torsional-angle changes to reduce the effect of load-estimation error on direction selection. The joint then moves through the sequence of compliant postures shown in Fig.~\ref{fig:joint-control}(g) and holds the new posture after the external force is removed. This experiment suggests that the load-adaptive structure of DANC can provide a basis for system-level force-interaction control in soft--rigid coupled manipulators.

\begin{figure*}[!t]
\centering
\includegraphics[width=\textwidth]{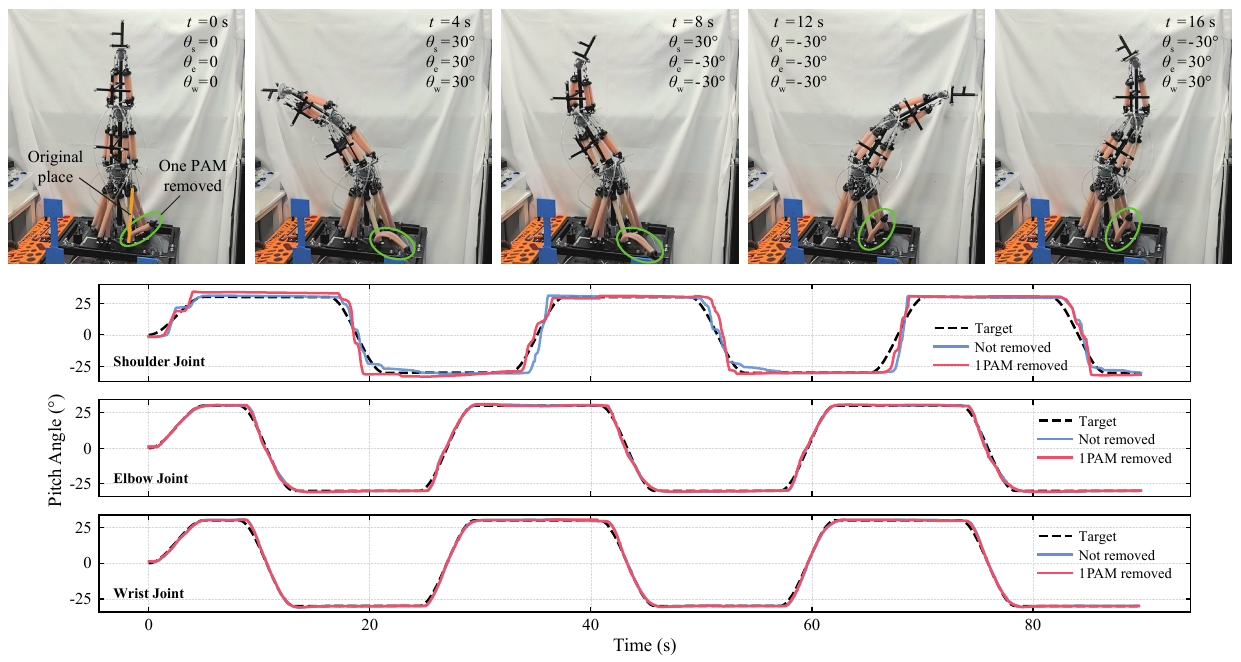}
\caption{Robustness evaluation after removing one PAM from the arm. The top row shows representative poses after PAM removal. The bottom plots compare the desired pitch trajectory with the measured shoulder, elbow, and wrist pitch angles before and after removal.}
\label{fig:arm-control-broken}
\end{figure*}

\subsection{Full-Arm Coordinated Motion}
The controller is then deployed on the complete 7-DOF arm. The commanded and measured shoulder, elbow, and wrist joint angles are mapped through the arm kinematics and PAM routing geometry to 22 PAM elongation references, which are tracked by the corresponding actuator-level DANC instances. Three motion-capture rigid-body markers are mounted on the links and end effector; together with the world frame $\{W\}$, they are used to measure the actual joint angles. Fig.~\ref{fig:arm-control-main} reports a continuous coordinated-motion trial: the bottom plots show the target and measured joint-angle trajectories, and the top two rows show five uniformly sampled physical snapshots with reconstructed poses and end-effector errors from the same trial.

The full-arm test is more difficult than the single-PAM and single-joint tests because the shoulder is never operating under a small load. Even without an external payload, the elbow module, wrist module, and the two distal links act as gravitational loads on the shoulder. Let $\mathcal{J}$ denote these distal components, with mass $m_j$ and center-of-mass position $\mathbf{r}_j$ relative to the shoulder center $\mathbf{r}_s$. The gravitational moment applied to the shoulder is
\begin{equation}
    \boldsymbol{\tau}_{s,g}
    = \sum_{j\in\mathcal{J}}
    \left(\mathbf{r}_j-\mathbf{r}_s\right)\times m_j\mathbf{g}.
\end{equation}
Because the arm is long, these moment arms can be large and change rapidly as the arm moves, producing a strongly posture-dependent equivalent load on the shoulder PAMs. This is the main reason the full-arm experiment is substantially harder than the isolated joint experiment.

In Fig.~\ref{fig:arm-control-main}, the measured elbow and wrist angles generally follow the target trajectories, while the shoulder angles show larger errors. Across the five sampled poses, the maximum end-effector position and orientation errors are 34.9 mm and $2.74^\circ$, respectively. The largest position error occurs in a highly bent intermediate posture with a large distal gravitational moment, which may contribute to the increased shoulder error. 
These results show that DANC can support system-level position control of the 7-DOF PAM-driven arm.

\subsection{PAM Removal}
Finally, Fig.~\ref{fig:arm-control-broken} evaluates the removal of one shoulder PAM that contributes directly to pitch actuation. The same full-arm pitch sequence is executed before and after removal, deliberately testing the motion direction affected by the missing actuator.
The removed PAM mainly affects the shoulder response, where a local loss of antagonistic actuation changes the available force balance and produces slightly different transients near direction reversals. The elbow and wrist traces remain close to the original response because their own PAM groups are still intact and the distal joints are controlled independently.

This test evaluates the passive redundancy of the actuation layout after a single local actuator loss. The controller is not explicitly re-optimized after the PAM is removed, and no failure-detection or pressure-redistribution algorithm is introduced. Nevertheless, the arm continues to execute the pitch sequence because the crossed antagonistic layout provides more actuation paths than the minimum required for the commanded motion, and DANC can absorb part of the changed force balance as an effective load shift.

\section{Discussion}
\label{sec:discussion}
The results support the central design choice of treating PAM regulation as both load-adaptive and history-dependent: the static characterization shows that pressure and axial load jointly determine elongation, while the dynamic and long-term tests show that the inverse pressure command also depends on quasi-static and rate-dependent hysteresis, pneumatic delay, creep, specimen variation, and aging drift. DANC separates these effects into components with specific physical roles---load-specific MLPs for the nonlinear inverse map, the elongation history for inertial effects and hysteretic memory, the preview for pneumatic-delay feedforward, the LEC MLP for fast load-change and disturbance response, the static compensator for residual steady-state error, and the stiffness-ratio correction for retraining-free specimen calibration---rather than an undifferentiated black-box correction. The LEC ablation further confirms that dynamic estimation and static compensation are complementary, with their fusion providing the best overall response to load changes.

The lower accuracy of the full-arm results reflects a combination of posture-dependent distal loading, parallel PAM load sharing, specimen-to-specimen mismatch, and cross-joint coupling. The shoulder errors in Fig.~\ref{fig:arm-control-main} should therefore be interpreted as system-level tracking errors accumulated from actuator regulation, pressure allocation, and coupled arm mechanics, rather than as single-PAM control errors alone.

The experiments illustrate the tradeoff behind treating $\hat m_t$ as a latent coordinate rather than a calibrated force: in the variable-load and calibration tests the estimate follows load changes at the trend level while its offset shifts so that the selected PIC load coordinate better reduces elongation error, and the drag-following test reuses the same estimate as a quasi-static interaction cue, indicating a possible signal for system-level force-interaction control.

The drift and PAM-removal experiments show retained tracking in the two tested cases: one PAM was retested after a 12-month interval, and the arm completed the tested pitch sequence after one PAM was removed. These results provide preliminary evidence of tolerance to the specific changes tested, rather than a general evaluation of long-term or fault robustness.

The cross-study comparison scales each response against a damping-matched second-order reference constructed from its effective mass and stiffness. DANC achieves the lowest time ratios among the cases for which these quantities are available or can be bounded, while the end-to-end ablation has a slower normalized response, larger overshoot and does not satisfy the 5\% settling criterion, supporting the dual-channel architecture. The comparison remains approximate because it does not remove differences in valves, supply pressure, tubing, sensing bandwidth, unreported fixture mass, or higher-order dynamics. It should therefore be interpreted as a plant-normalized experimental comparison rather than an identical-hardware ranking.

Several limitations remain. The main limitation is that the present implementation does not yet make whole-arm control decisions over redundant tensions; instead, the desired motion is resolved kinematically into individual PAM elongation commands, and DANC regulates each PAM locally. Future work will focus on arm-level allocation and interaction decisions, including redundant pressure coordination and contact-aware control. The current experiments also rely on OptiTrack-based shape sensing. Embedded sensing will be needed for operation outside the calibrated motion-capture volume.

\section{Conclusion}
This paper presented a 7-DOF anthropomorphic pneumatic manipulator driven by 22 custom-fabricated multi-fiber PAMs and DANC, a load-adaptive actuator-level controller for PAM regulation. Characterization of the PAMs and joints first identified the main control challenges: coupled pressure--load nonlinearity, motion-history effects, specimen variation, long-term drift, and posture-dependent joint torque and stiffness.

To address these effects, DANC combines history-aware pressure inference, online load-coordinate estimation, and rapid stiffness-ratio calibration. Controller ablations support the dual-channel architecture and the complementary roles of dynamic estimation and static compensation. Experiments evaluated the approach from actuator to arm scale: DANC achieved the lowest plant-normalized rise- and settling-time ratios among the published single-PAM cases included in the comparison, a single PAM tracked trajectories under changing loads, the 3-DOF spherical joint tracked star-shaped paths with a best RMSE of 6.7 mm, and the complete arm executed coordinated 7-DOF motion with maximum sampled end-effector errors of 34.9 mm and $2.74^\circ$. 
The 12-month repeat test on one PAM and a single-PAM-removal trial further showed retained motion under those specific test conditions.

These results indicate that interpretable, load-adaptive PAM regulation can be composed from individual soft actuators to an anthropomorphic soft--rigid arm. Future work will move from local PAM regulation toward whole-arm redundant tension allocation, embedded shape sensing, and contact-aware interaction control.

\bibliographystyle{IEEEtran}
\bibliography{refs}

\end{document}